\RequirePackage{fix-cm}
\documentclass[smallextended]{svjour3}       % onecolumn (second format)
\smartqed  % flush right qed marks, e.g. at end of proof
\usepackage{graphicx}
\graphicspath{{./figs/}}

\usepackage{amsmath}
\usepackage{amssymb}
\newcommand{\LSE}{\mathrm{LSE}}
\newcommand{\Rwk}{R^k_{\overline{w}}}
\newcommand{\Rzk}{R^k_{\overline{z}}}
\newcommand{\Rxk}{R^k_{\overline{x}}}
\newcommand{\Dk}{V^k}

\usepackage{pgfplots}%\pgfplotsset{compat=1.4}
\usepackage{caption}
\usepackage{subcaption}
\usepackage{tabularx}

\usepackage[numbers,sort&compress]{natbib}

\usepackage{etoolbox}
\usepackage{mathtools}

\newcommand\numberthis{\addtocounter{equation}{1}\tag{\theequation}}

\DeclarePairedDelimiter{\dotp}{\langle}{\rangle}
 \DeclarePairedDelimiter{\ceil}{\lceil}{\rceil}

\usepackage{booktabs}

\usepackage{mathtools}
\usepackage[normalem]{ulem} % to use \sout
\newcommand{\remove}[1]{{}}

\newcommand{\cB}{{\mathcal{B}}}

\newcommand{\cF}{{\mathcal{F}}}

\newcommand{\cH}{{\mathcal{H}}}
\newcommand{\cI}{{\mathcal{I}}}

\newcommand{\cO}{{\mathcal{O}}}

\newcommand{\RR}{\mathbb{R}}
\newcommand{\NN}{\mathbb{N}}

\newcommand{\EE}{\mathbb{E}}

\newcommand{\dom}{{\mathrm{dom}}} % domain
\newcommand{\prox}{\mathbf{prox}}

\DeclareMathOperator*{\minimize}{minimize}
\DeclareMathOperator*{\argmin}{argmin}

\DeclareMathOperator*{\as}{a.s.} 

\newcommand{\bc}{\begin{center}}
\newcommand{\ec}{\end{center}}

\newcommand{\bdm}{\begin{displaymath}}
\newcommand{\edm}{\end{displaymath}}

\newcommand{\beq}{\begin{equation}}
\newcommand{\eeq}{\end{equation}}

\newcommand{\bfl}{\begin{flushleft}}
\newcommand{\efl}{\end{flushleft}}

\newcommand{\bt}{\begin{tabbing}}
\newcommand{\et}{\end{tabbing}}

\newcommand{\beqn}{\begin{align}}
\newcommand{\eeqn}{\end{align}}

\newcommand{\beqs}{\begin{align*}} % no equation numbers
\newcommand{\eeqs}{\end{align*}}  % no equation numbers

\newtheorem{assumption}{Assumption}

\usepackage{algorithm,algpseudocode}

\begin{document}

\title{A SMART Stochastic Algorithm for Nonconvex Optimization with Applications to Robust Machine Learning%\thanks{Grants or other notes
%about the article that should go on the front page should be
%placed here. General acknowledgments should be placed at the end of the article.}
}
%\subtitle{Do you have a subtitle?\\ If so, write it here}

\titlerunning{A SMART Stochastic Algorithm for Nonconvex Optimization}        % if too long for running head

\author{Aleksandr Aravkin\thanks{This work was funded by the Washington Research Foundation Data Science Professorship.}        \and
        Damek Davis\thanks{This material is based upon work supported by the National Science Foundation under Award No. 1502405.} %etc.
}

%\authorrunning{Short form of author list} % if too long for running head

\institute{A. Aravkin \at
               Department of Applied Mathematics\\
  University of Washington\\
  Seattle, WA 98195-4322, USA  \\            
  \email{saravkin@uw.edu}           %  \\
%             \emph{Present address:} of F. Author  %  if needed
           \and
           D. Davis \at
              School of Operations Research and Information Engineering\\
Cornell University\\
 Ithaca, NY 14850, USA\\
 \email{dsd95@cornell.edu}
}

\date{Received: date / Accepted: date}
% The correct dates will be entered by the editor

\maketitle

\begin{abstract}
In this paper, we show how to transform any optimization problem that arises from fitting a machine learning model into one that (1) detects and removes contaminated data from the training set while (2) simultaneously fitting the trimmed model on the uncontaminated data that remains. To solve the resulting nonconvex optimization problem, we introduce a fast stochastic proximal-gradient algorithm that incorporates prior knowledge through nonsmooth regularization. For datasets of size $n$, our approach requires $O(n^{2/3}/\varepsilon)$ gradient evaluations to reach $\varepsilon$-accuracy and, when a certain error bound holds, the complexity improves to $O(\kappa n^{2/3}\log(1/\varepsilon))$. These rates are $n^{1/3}$ times better than those achieved by typical, full gradient methods.
% \PACS{PACS code1 \and PACS code2 \and more}
% \subclass{MSC code1 \and MSC code2 \and more}
\end{abstract}

\section{Introduction}

Potential outliers in datasets can be identified in several ways. For low-dimensional models, scatter plots, box plots, and histograms can be used to visually identify points that deviate from modeling assumptions. For higher-dimensional data, several tests involving order statistics exist (so called L-estimators~\citep{mar}), such as the three-sigma rule for Gaussian data, or trimming strategies for disregarding points that are furthest away from the mean. After potential outliers are removed from a dataset, models are fit on the remaining data. After fitting the model, potential outliers are again identified and removed and another model is fit~\citep{ruppert1980trimmed}. This process can repeat indefinitely, until no points are left in the  dataset.

Identifying outliers using a fitted model can be problematic, since outliers  affect the fit.  
Robust loss functions are often used to estimate model parameters from potentially contaminated data, without any \textit{a priori} outlier removal or pre-processing. Examples include the $\ell_1$, huber, and 
Student's t losses, all of which attempt to minimize the {\it influence} of observations that deviate from modeling assumptions~\citep{hub,Lange1989}. After fitting a model using a robust loss, potential outliers can be identified by sorting the loss applied to individual observations. Observations with higher loss are considered more likely to be outliers. 

Another approach, called trimmed estimation, couples explicit outlier identification and removal with model fitting. 
Given a set of $n$ training examples, typical model fitting, i.e., M-estimation, solves  
\begin{align*}
\minimize_x \sum_{i=1}^n f_i(x), 
\end{align*}
where each $f_i$ represents the loss associated with the $i$th training example. 
In contrast, trimmed M-estimators couple this already difficult, potentially nonconvex, optimization problem with explicit outlier removal
\begin{align}\label{eq:trimmed}
\minimize_x \sum_{i=1}^h f_{i:n}(x),
\end{align}
where $f_{1:n}(x) \leq \cdots \leq f_{h:n}(x)$ are the first $h$ order statistics of the objective values. If loss $f_i$ is the log likelihood of the $i$th observed sample, then trimming attempts to jointly fit a probabilistic model while simultaneously eliminating the influence of all low likelihood observations. 

Trimmed M-estimators were initially introduced by~\cite{rousseeuw1985multivariate} in the context of least-squares regression. The author's original motivation was to develop linear regression estimators that have a high breakdown point (in this case 50\%) and good statistical efficiency (in this case $n^{-1/2}$)\footnote{Breakdown refers to the percentage of outlying points which can be added to a dataset before the resulting M-estimator can change in an unbounded way. Here, outliers can affect both the outcomes and training data (features).}. These Least Trimmed Squares (LTS) estimators were proposed as a higher efficiency alternative to Least Median Squares (LMS) estimators~\citep{rousseeuw1984least}, which replace the sum in~\eqref{eq:trimmed} by a median. For a number of years, the difficulty of efficiently optimizing LTS problems limited their application. The problem is difficult because

\begin{quote}
even if all losses $f_i$ are smooth and convex,~\eqref{eq:trimmed} is, in general, nonsmooth and nonconvex.%\footnote{For example, with $f_1(x) = (1+x)^2$, $f_2(x) = (1-x)^2$, and $h=1$, the nonconvex, trimmed objective $\min\{(1+x)^2, (1-x)^2\}$ is nonsmooth at the origin.}
\end{quote}

Nevertheless, several approaches for finding LTS and other trimmed M-estimators have been developed. The authors of \cite{rousseeuw2006computing}  developed the FAST-LTS algorithm, which was able to find LTS estimators faster than existing algorithms for LMS estimations. Later, \cite{mount2014least} introduced an exact algorithm for computing LTS, which suffered from exponential complexity in higher dimensional problems. Generalizing the approach developed in \cite{rousseeuw2006computing}, \cite{neykov2003breakdown}  developed the FAST-TLE method, which replaces the least squares terms in the LTS formulation with log-likelihoods of generalized linear models. In a different direction, \cite{alfons2013sparse} proposed a sparse variant of the Fast-LTS algorithm for L1-regularized LTS estimation. Further work in \citep{yang2015robust,yang2016high}  proposed algorithms for graphical lasso and regularized trimming of convex losses. 

With the exception of \cite{mount2014least,yang2015robust,yang2016high}, each of the proposed algorithms above are variants of the alternating minimization algorithm. The algorithms in \cite{yang2015robust,yang2016high} mixed alternating minimization and proximal-gradient steps. The algorithm of \cite{mount2014least} is combinatorial in nature, but has exponential complexity.

There are two drawbacks to trimming algorithms based on alternating minimization. First, they are greedy algorithms, which do not always work well for nonconvex problems; and second, they require, at every iteration, solving a large optimization problem typically involving more than 50\% of the dataset.\footnote{For example, \cite{alfons2013sparse} requires solving a full LASSO problem at each iteration. And although the algorithm of~\citep{yang2016high} requires only one pass over the dataset at each iteration, this is still problematic for large datasets.} The first drawback is well-known in the optimization community, while the second is motivation for introducing stochastic gradient approaches for trimming. 

At first glance, the standard stochastic gradient (SG) method appears to be the natural algorithm for solving~\eqref{eq:trimmed}. However,~\eqref{eq:trimmed} is nonsmooth and nonconvex, so there are, as of yet, no known convergence rate guarantees for SG applied to~\eqref{eq:trimmed}. In this paper we develop a variance-reduced stochastic gradient algorithm with convergence rate guarantees.

\subsection{Contributions}

\paragraph{Fully Nonconvex Problem Class.} 
Our new algorithm extends the Stochastic Monotone Aggregated Root-Finding (SMART) algorithm~\citep{davis2016smart} to the nonsmooth, nonconvex trimming problem. To keep with tradition, we call this algorithm SMART. It is the first variance-reduced stochastic gradient algorithm for fully nonconvex optimization (our losses and our regularizers are nonconvex). It also applies to much more general problems than~\eqref{eq:trimmed}. We consider the following class: 
\begin{equation}
\label{eq:trimmedformulation}
\minimize_{w \in\RR^{n}, \; x \in \cH} \left\{\frac{1}{n}\sum_{i=1}^n w_i  f_i(x) + r_1(w) + r_2(x)\right\},
\end{equation}
where each $f_i$ is $C^1$ and $r_1$ and $r_2$ are lower semincontinuous (potentially nonconvex) functions. This more general problem class recovers~\eqref{eq:trimmed}: simply let $r_1 : \RR^n \rightarrow [0, \infty]$ be the indicator function of the capped simplex
$$
\Delta_h := \left\{(w_1, \ldots, w_n) \mid w \in [0, 1], \sum_{i=1}^n w_i = h\right\},
$$
and minimize jointly over $w$ and $x$.

\paragraph{Better Dependence on Lipschitz Constants.} It is possible to \textit{apply} the proximal gradient algorithm to this problem\footnote{For example, the pioneering work of \cite{Attouch2013} proved that the proximal gradient algorithm converges under extremely general conditions.} 
but its convergence is not guaranteed without taking very small stepsizes. This restriction arises because the standard sufficient condition for guaranteeing the convergence of the proximal gradient method requires using a stepsize that is proportional to the inverse of the Lipschitz constant of the gradient of the smooth function $G(w,x) = w_i  f_i(x) $, which is not globally Lipschitz: $\nabla G(w, x) = (f_i(x), w_i\nabla f_i(x))$. Even for least squares problems, the local Lipschitz constant of $\nabla G(w, x)$ grows with $\|x\|$ and $\|w\|$. This issue likewise prevents our using the ProxSAGA and ProxSVRG~\citep{reddi2016fast}.

\paragraph{Convergence Rates that Scale with $n^{2/3}$.} A good alternative to the proximal-gradient method is called the Proximal Alternating Linearized Minimization (PALM) method~\citep{bolte2014proximal} (see Section~\ref{sec:algorithm}), which allows for stepsizes that only scale inversely with $\|w\|$ and the Lipschitz constants of $\nabla f_i$. The convergence rate of this algorithm was analyzed in the fully nonconvex case in~\cite[Theorem 5.4]{davis2016asynchronous}, where it was shown that an $\varepsilon$-stationary point (see Section~\ref{sec:sublinear}) could be found within $O(1/\varepsilon)$ iterations. Thus, in total PALM finds $\varepsilon$-stationary points using $O(n/\varepsilon)$ gradients.

SMART scales better than PALM and other competing methods by a factor of $n^{1/3}$. In particular, without any regularity assumptions
\begin{quote}
 SMART finds an $\varepsilon$-stationary point with $O(n + n^{2/3}/\varepsilon)$ gradient evaluations
\end{quote}
(see Corollaries~\ref{cor:sublinear} and~\ref{cor:sublinear2}). This matches the complexity of ProxSAGA/ProxSVRG~\citep{reddi2016fast}, which only apply to the special case of problem~\eqref{eq:trimmedformulation} considered in Section~\ref{sec:prox_saga}.

When a certain error bound holds (see~\eqref{eq:error_bound}), 
\begin{quote}
SMART finds an $\varepsilon$-stationary point with $O\left(n + \kappa n^{2/3}\log\left(1/\varepsilon\right)\right)$ gradient evaluations,
\end{quote}
where $\kappa$ is akin to a condition number of~\eqref{eq:trimmedformulation} (see Corollaries~\ref{cor:linear_saga} and~\ref{cor:linear_SVRG}). In contrast, ProxSAGA and ProxSVRG~\citep{reddi2016fast}, which only apply to the special case of problem~\eqref{eq:trimmedformulation} considered in Section~\ref{sec:prox_saga}, both require $O((n + \kappa n^{2/3})\log\left(1/\varepsilon\right))$ gradient evaluations to reach accuracy $\varepsilon$.

\paragraph{Organization.}
We present algorithms related to SMART in Sections~\ref{sec:prox_saga} and~\ref{sec:partial}. We also present several theoretical guarantees for SMART in Section~\ref{sec:theory}. In Section~\ref{sec:numerics}, we perform three trimming experiments.  We present robust digit recognition for the {\bf mnist} dataset,  introduce trimmed Principal Component Analysis to determine the quality of judges in the {\bf USJudges} dataset, 
and apply SMART to find a homography between two images using interest point matching. Proofs of the main theorems are presented in the appendices.

\subsection{Notation}
\label{sec:notation}
In  Problem~\eqref{eq:trimmedformulation}  the variable $x$ is an element of a finite dimensional Euclidean space $\cH$; each function $f_i : \cH \rightarrow \RR$ is $C^1$, each gradient $\nabla f_i$ is $L$-Lipschitz continuous;  both functions $r_1 : \RR^n \rightarrow (-\infty, \infty]$ and $r_2 : \cH \rightarrow (-\infty, \infty]$ are proper and lower-semicontinuous. We assume that the point-to-set proximal mapping $\prox_{\gamma r_j} : \cH \rightarrow 2^\cH$
$
x \mapsto  \argmin_{x' \in \cH} \left\{r_j(x') + (1/(2\eta))\|x' - x\|^2\right\}
$
is always nonempty for every $\eta$ small enough, say for $\eta  < \delta_{r_1}$ if $j = 1$ and for $\eta < \delta_{r_2}$ if $j = 2$. 

We work with an underlying probability space denoted by $(\Omega, \cF, P)$, and we assume that the space $\cH$ is equipped with Borel $\sigma$-algebra $\cB$. An $\cH$-valued random variable is a measurable map $X : (\Omega, \cF) \rightarrow (\cH, \cB)$. We always let $\sigma(X) \subseteq \cF$ denote the sub $\sigma$-algebra generated by a random variable $X$. We use the shorthand $\as$ to denote almost sure convergence of a sequence of random variables. By our assumptions on  $r_1$ and $r_2$, for $j \in \{1, 2\}$ there exists measurable mappings $\zeta_j : \cH_j \times (0, \delta_{r_j})\rightarrow \cH_j$ such that $\zeta_j(x, \gamma) \in  \prox_{\gamma r_j}(x)$ for all $(x, \gamma) \in \cH_j \times (0, \delta_{r_j})$, where $\cH_1 = \RR^n$ and $\cH_2 = \cH$ \citep{RTRW}. For the rest of the paper, we let $x^+ = \prox_{\gamma r_j}(x)$ mean that $x^+ = \zeta_j(x, \gamma)$. 

We use the notation 
$$
F(w, x) = \frac{1}{n}\sum_{i=1}^n w_i  f_i(x) + r_1(w) + r_2(x)
$$
throughout the paper and assume  $(w^\ast, x^\ast) \in \argmin_{x\in \cH, w \in \RR^n} F(w, z)$ exists.

We assume that $\dom(r_1)$ is bounded: there exists $B_i > 0$ such that for all $ w\in \dom(r_1)$, we have $|w_i| \leq B_i$.

\vspace{-20pt}\section{Algorithm}
\label{sec:algorithm}
To find a stationary point of~\eqref{eq:trimmedformulation}, our algorithm iteratively updates a state vector $(w^k, x^k) \in \RR^n\times \cH$. The algorithm is designed so that $(w^k, x^k) $ will not only be close to a stationary point after just a few iterations, but so that the average computational complexity of obtaining $(w^{k+1}, x^{k+1}) $ from $(w^{k}, x^k)$ will be small. 
These competing objectives can both be achieved simultaneously by combining ideas from the \emph{Proximal Alternating Linearized Minimization} (PALM) method~\citep{bolte2014proximal}, which obtains $(w^{k+1}, x^{k+1}) $ from $(w^k, x^k) $ via 
\begin{align*}
w^{k+1} &:= \prox_{\tau r_1}\left(w^k - \frac{\tau}{n}(f_1(x^k), \ldots, f_n(x^k))\right);\\
x^{k+1} &:=  \prox_{\gamma r_2}\left(x^k - \frac{\gamma}{n}\sum_{i=1}^n w_i^{k+1}\nabla f_i(x^k)\right),
\end{align*}
and the \emph{partially stochastic proximal-gradient} (PSPG) method, which obtains $(w^{k+1}, x^{k+1})$ from $(w^{k}, x^{k})$ via 
\begin{align*}
w^{k+1} &:= \prox_{\tau r_1}\left(w^k - \frac{\tau}{n}(f_1(x^k), \ldots, f_n(x^k))\right);\\
x^{k+1} &:=  \prox_{\gamma r_2}\left(x^k - \frac{\gamma_k}{n} w_{i_k}^k \nabla f_{i_k}(x^k)  \right),
\end{align*}
where $i_k \in \{1, \ldots, n\}$ is randomly sampled and $\gamma_k \rightarrow 0$ as $k \rightarrow \infty$. 

PALM takes few iterations to obtain near stationary $(w^k, x^k)$ ($\varepsilon$ accuracy obtained after $O(1/\varepsilon)$ iterations), but for each $k$ it computes the full gradient $ n^{-1}\sum_{i=1}^n w_i^k\nabla f_i(x^k)$, which can be costly. On the other hand, PSPG takes many iterations to obtain near stationary $(w^k, x^k)$, but for each $k$ it only computes a single gradient $w_{i_k}^k\nabla f_{i_k}(x^k)$, which can be done quickly. But for nonconvex problems, \textbf{there is no known rate of convergence for PSPG} (unless minibatches of stochastic gradients of increasing size are used~\citep{ghadimi2016mini,davis2016sound}). Even in the relatively simple case where $f_i$, $r_1$, and $r_2$ are convex, there is still a nonconvex coupling between $w_i$ and $f_i$ and, hence, no known rate of convergence for PSPG. 

By reducing the variance of the stochastic gradient estimator $w_{i_k}\nabla f_{i_k}$, we create a fast algorithm, which we call SMART, 
that combines the PALM and PSPG updates and obtains an $\varepsilon$ accuracy solution after $O(1/\varepsilon)$ steps. As in PSPG, SMART typically evaluates a single gradient $\nabla f_{i_k}$ (or a small batch) at one or two points per iteration. But unlike PSPG, SMART on average only evaluates all the function values $(f_1(x^k), \ldots, f_n(x^k))$ once per every $t$ iterations, where $t$ is user defined.

\subsection{Implementation and Features } 

\begin{algorithm}
  \caption{SMART for~\eqref{eq:trimmedformulation}}
  \label{alg:3split}
  \begin{algorithmic}[1]
     \State Choose $\gamma  < \delta_{r_1}$; \; $\tau < \delta_{r_2}$; \; $(w^0, x^0) \in \dom(r_1)\times\dom(r_2)$; \; $y_{i}^0 = \nabla f_i(x^0)^Tw_i^0$.
     \For{$k = 0, 1, \ldots $}
     \State Sample $I_k \subseteq \{1, \ldots, n\}$; $j_k \in \{1,2\}$; $ D_k \subseteq  \{1, \ldots, n\}$;
     \If{$j_k = 1$} 
     \State $w^{k+1} \gets \prox_{\tau r_1}\left(w^k - \frac{\tau}{n}(f_1(x^k), \ldots, f_n(x^k))\right)$;
     \State $x^{k+1} \gets x^k$;
     \For{$i = 1, \ldots, n$}
     \State $y_i^{k+1} \gets  w_i^{k+1} \nabla f_i(x^{k})$;
     \EndFor
    \Else
     \State $w^{k+1} \gets w^k$;
     \State $x^{k+1} \gets \prox_{\gamma r_2}\left(x^k - \gamma\left( \frac{1}{b}\sum_{i \in I_k}(w_{i}^k \nabla f_{i}(x^k) - y_{i}^k) + \frac{1}{n} \sum_{i=1}^n y_{i}^k\right)\right)$;
     \For{$i \in D_k$}
     \State $y_i^{k+1} \gets   w_i^{k}\nabla f_i(x^{k}) $;
     \EndFor
    \EndIf    
    \EndFor
  \end{algorithmic}
\end{algorithm}

\paragraph{Incremental Gradients and Minibatches.}

Rather than evaluating a full gradient $\nabla f = n^{-1}\sum_{i=1}^n \nabla f_i$ at each iteration, we instead sample $b$ elements uniformly at random with replacement and denote this collection by $I_k \subseteq \{1, \ldots, n\}$; then we only evaluate $\nabla f_{i}$ for $i \in I_k$. We assume $\{I_k\}_{k \in \NN}$ is IID.

\paragraph{Block Coordinates Updates.}

At every iteration we sample a coordinate $j_k \subseteq \{1,2\}$ that indicates whether $w^k$ is modified ($j_k = 1$) or whether $x^k$ is modified ($j_k = 2$) to obtain $(w^{k+1}, x^{k+1})$. We assume that $\{j_k\}_{k \in \NN}$ is IID and the variables $I_k$ and $j_k$ are independent. We let 
\begin{align*}
q &:= P(j_k = 1) > 0 ,  & q' = P(j_k = 2) > 0.
\end{align*}

\paragraph{Dual Variables and Dual Updates.} 

For each index $i \in \{1, \ldots, n\}$, we maintain a sequence of \emph{dual variables}, denoted by $y_i^k\in \cH$. The dual variables are always parametrically defined: $y_{i}^k = \phi_{i1}^k  \nabla f_i(\phi_{i2}^k)  $ for old iterates $(\phi_{i1}^k, \phi_{i2}^k) \in \{(w_i^l, x_i^l)\}_{l < k}$. The sum $n^{-1}\sum_{i=1}^n y_{i}^k$ approximates the gradient $n^{-1}\sum_{i=1}^n w_i^k\nabla f_i(x^k)$ and is used in the following stochastic estimator of the sum, which has smaller variance than the SG estimator $w_{i}^k \nabla f_{i}(x^k)$:
\begin{align}\label{eq:stochasticestimator}
\frac{1}{b} \sum_{i \in I_k} (w_{i}^{k} \nabla f_{i}(x^k) - y_{i}^k) + \frac{1}{n} \sum_{i=1}^n y_{i}^k.
\end{align}

The dual variables need not be recomputed at every iteration, so $\phi_{i}^k$ can be quite a stale estimate of $x^k$. We introduce the set-valued random variable and probability
\begin{align*}
D_k \subseteq  \{1, \ldots, n\}; && \text{and} && \rho_{i} := P(i\in D_k),
\end{align*}
which control whether the $i$th dual variable is updated at iteration $k$: 
\begin{align*}
 y_{i}^{k+1} = 
\begin{cases}
w_i^k \nabla f_i(x^k) & \text{if $i\in D_k$;} \\
y_{i}^k & \text{otherwise.}
\end{cases}
\end{align*}
We assume that $\{D_k \}_{k \in \NN}$ is IID and that $D_k$ is independent from $j_k$, but we do not assume that $D_k$ is independent from $I_k$.

 \vspace{-9pt}\subsection{Connection to ProxSAGA and ProxSVRG.}\label{sec:prox_saga}

Our main goal is to use the regularizer $r_1$ to trim statistical models, but we can turn off trimming  
by choosing $r_1$ to be the convex, $\{0, \infty\}$-valued indicator that forces all weights $w_i$ to be $1$. In this case, we recover and extend the ProxSAGA algorithm, introduced by~\cite{defazio2014saga} and recently analyzed for nonconvex problems by~\cite{reddi2016fast}, by letting $D_k$ be a set consisting of $b$ elements of $\{1, \ldots, n\}$, sampled uniformly at random with replacement, and by letting $q = 0$. In terms of implementation,  we never perform a $w$ or a full gradient update, but at every iteration we update the dual variable $y_i^k$ for  $i \in D_k$. \textbf{Our work extends the work by~\cite{reddi2016fast} by allowing nonconvex regualizers $r_2$}, whereas~\cite{reddi2016fast} requires $r_2$ to be convex.

We also recover a variant of ProxSVRG, introduced by~\cite{proxsvrg} and recently analyzed for nonconvex problems analyzed by~\cite{reddi2016fast}, by setting $D_k = \emptyset$ and $q = 1/t$, where $t > 1$ is the average number of iterations we wish to perform before recomputing a full gradient. Although it appears that the $w$ step requires a computation of the function values $(f_1(x^k), \ldots, f_n(x^k))$, it does not because $w_i^k \equiv 1$. As in the ProxSAGA case, our work extends~\cite{reddi2016fast} by allowing nonconvex regularizers $r_2$.

\vspace{-9pt}\subsection{Connection to Partial Minimization and Randomized Coordinate Descent}\label{sec:partial}

With appropriate choices of the random variables $j_k$, $I_k$, and $D_k$, we recover randomized variants
of PALM~\citep{bolte2014proximal} and the full gradient method of~\cite{aravkin2016variable}. The key is to choose $I_k = D_k \equiv \{1, \ldots, n\}$, so that all dual variables are constantly updated, and $q := P(j_k = 1) = 1/2$. Then, our stochastic estimator~\eqref{eq:stochasticestimator} is equal to the full gradient: $n^{-1} \sum_{i=1}^n w_i^k \nabla_i f_i(x^k).$ For fixed $\tau$, we get a randomized variant of the algorithm of~\cite{bolte2014proximal}. For $\tau \rightarrow \infty$, we get a method similar to that of~\cite{aravkin2016variable}, except that we allow nonconvex regularizers. When $r_2$ is convex, $\prox_{\tau r_2}(w)$ converges to an element of $\argmin \{r_2(w)\}$~\cite[Theorem 23.44]{bauschke2011convex}; in the general case $r_2$ need only be prox bounded, so $\prox_{\tau r_2}$ may not even be defined for large $\tau$.

\section{Convergence Theory}

Our convergence rates are organized in Table~\ref{table:rates}. We separate our sublinear and linear convergence rate results into Section~\ref{sec:sublinear} and~\ref{sec:linear}, respectively.

\label{sec:theory}

\vspace{-20pt}\subsection{Sublinear Rates}\label{sec:sublinear}

\paragraph{$\varepsilon$-Stationary Points.}
For all $k \in \NN$, we define $\overline{x}^{k+1}\in \cH$ and $\overline{w}^{k+1} \in \RR^{n}$ by: 
\begin{align*}
\overline{w}^{k+1} &:= \prox_{\tau r_1}\left(w^k - \frac{\tau}{n}(f_1(x^k), \ldots, f_n(x^k))\right)\\
\overline{x}^{k+1} &:= \prox_{(\gamma/\eta) r_2}\left(x^k -\frac{\gamma}{\eta n}\sum_{i=1}^n w_i^k\nabla f_i(x^k)\right).
\end{align*}
SMART never actually computes $\overline{x}^{k+1}$; it is only used in the analysis of the algorithm. Its existence shows that a nearby, nearly stationary point can be obtained with $n$ gradient evaluations. For our analysis, it is crucial that $\eta$ be a constant greater than 1, i.e., we must shorten the steplength in order to measure stationarity. 

We measure convergence of $(w^k, x^k)$ by bounding the normalized step sizes
\begin{align*}
\frac{1}{\tau}\left(w^k - \overline{w}^{k+1}\right) &\in \frac{1}{n}(f_1(x^k), \ldots, f_n(x^k)) + \partial_L r_1(\overline{w}^{k+1}); \\
\frac{\eta}{\gamma}\left(x^k - \overline{x}^{k+1}\right) &\in \frac{1}{n}\sum_{i=1}^n w_i^k  \nabla f_i(x^k)+ \partial_L r_2(\overline{x}^{k+1}),
\end{align*}
where $\partial_L r_j$ denotes the \emph{limiting subdifferential} of $r_j$ \citep[Definition 8.3]{RTRW}. It is common to compute bounds on the square of these step lengths, although it is perhaps misleading to do so. To make it easy to compare our results with the current literature, we also bound the squared steplengths Theorem~\ref{thm:smartconverges}.

Using the Lipschitz continuity of $\nabla f_i$ and the local Lipschitz continuity of $f_i$, these bounds easily translate bounds on
$
\mathrm{dist}\left(0, \partial_L F(\overline{w}^{k+1}, \overline{x}^{k+1})\right).
$
We omit this straightforward derivation.

\begin{table}
\parbox{\linewidth}{
\centering
\begin{tabular}{lllllll|l}
 \toprule
\textbf{Algorithm}&\textbf{GradEvals}& \textbf{FunEvals}&\textbf{$\prox_{\tau r_1} $Evals}&\textbf{$\prox_{\gamma r_2} $Evals} \\\cmidrule{1-5}
SMART(SAGA) & $O(n + \frac{n^{2/3}}{\varepsilon})$&$O(\frac{1}{\varepsilon})$& $O(\frac{1}{n\varepsilon})$&$O(\frac{1}{\varepsilon})$\\%\hline
SMART(SAGA$+$\eqref{eq:error_bound}) &$O(n + \kappa n^{2/3}\log(\frac{1}{\varepsilon}))$&$O(\kappa\log(\frac{1}{\varepsilon}))$& $O(\kappa\log(\frac{1}{\varepsilon}))$ &$O(\frac{\kappa}{n}\log(\frac{1}{\varepsilon}))$\\%\hline
SMART(SVRG) &$O(n + \frac{n^{2/3}}{\varepsilon})$&$O(\frac{n^{2/3}}{\varepsilon})$&$O(\frac{1}{n^{1/3}\varepsilon})$&$O(\frac{1}{\varepsilon})$\\%\hline
SMART(SVRG$+$\eqref{eq:error_bound}) &$O(n + \kappa n^{2/3}\log(\frac{1}{\varepsilon}))$&$O(\kappa n^{2/3}\log(\frac{1}{\varepsilon}))$&$O(\frac{\kappa}{n^{1/3}}\log(\frac{1}{\varepsilon}))$&$O(\kappa\log(\frac{1}{\varepsilon}))$\\%\hline
PALM & $O(\frac{n}{\varepsilon})$&$O(\frac{n}{\varepsilon})$& $O(\frac{1}{\varepsilon})$&$O(\frac{1}{\varepsilon})$\\%\hline
PALM($+$\eqref{eq:error_bound}) &$O(\kappa n\log(\frac{1}{\varepsilon}))$&$O(\kappa n\log(\frac{1}{\varepsilon}))$&$O(\kappa\log(\frac{1}{\varepsilon}))$&$O(\kappa\log(\frac{1}{\varepsilon}))$\\
\bottomrule
\end{tabular}
\caption{\label{table:rates} Convergence rates of SMART and PALM in terms of number of operations needed to achieve accuracy $\varepsilon$. The constant $\kappa$ is defined in Section~\ref{sec:linear}.  The rates for SMART are proved in Corollaries~\ref{cor:sublinear},~\ref{cor:sublinear2},~\ref{cor:linear_saga}, and~\ref{cor:linear_SVRG}. The rates for PALM can be determined (with some effort) from the proofs in~\citep{davis2016asynchronous}. Alternatively, the rates for PALM may be derived from Theorems~\ref{thm:smartconverges} and~\ref{thm:convergencerate} by using the randomized variant of PALM discussed in Section~\ref{sec:partial}.}
}
\end{table}

\paragraph{Independence of Algorithm History and Sampling}

The SMART algorithm generates a sequence of random variables $\{(w^k, x^k)\}_{k \in \NN}$. Throughout the algorithm, we make the standard assumption that
\begin{assumption}\label{assumption:independence}
The $\sigma$-algebra generated by the history of SMART, denoted by $\cF_k = \sigma((w^0, x^0), \ldots, (w^k, x^k)),$  is independent of the $\sigma$-algebra $\cI_k = \sigma((I_k, j_k, D_k))$.
 \end{assumption}

SMART converges, provided we choose $\gamma$ properly. In measuring convergence, we introduce a particular $\eta > 0$ (which depends on a user defined constant $\epsilon_0 \in (0, 1)$):
\begin{align}\label{eq:eta_definition}
\eta &= 2+ 4\gamma\left[\sqrt{\frac{1}{n}\sum_{i=1}^n   \frac{q' (1+\epsilon_0)(B_iL)^2}{2b\left(1 - \sqrt{q'(1-\rho_{i})}\right)^2}} + \frac{4L}{n}\sum_{i=1}^n B_i\right].
\end{align}
This constant is key for showing that Algorithm~\ref{alg:3split} converges with nonconvex regularizers $r_1$ and $r_2$. We place the proof in Appendix~\ref{app:proof}.

\begin{theorem}[SMART Converges]\label{thm:smartconverges}
Suppose $\{(w^k, x^k)\}_{k \in \NN}$ is generated by Algorithm~\ref{alg:3split} and that Assumption~\ref{assumption:independence} holds. Let $\epsilon_0 \in (0, 1)$ and let $\eta$ be defined as in~\eqref{eq:eta_definition}. Then, if 
\begin{align*}
\gamma \leq \frac{1}{4L\sqrt{\frac{1}{n}\sum_{i=1}^n   \frac{q'(1+\epsilon_0) B_i^2}{2b\left(1 - \sqrt{q'(1-\rho_{i})}\right)^2}} + \frac{L}{n}\sum_{i=1}^n B_i  },
\end{align*}
the following hold:
\begin{enumerate}
\item \label{thm:smartconverges:item:decreaseinobjective}\textbf{Objective Decrease.} The limit $\lim_{k \rightarrow \infty} F(w^k, x^k)$ exists almost surely and for all $k \in \NN$, we have 
\begin{align*}
&\EE \left[ F(w^{k+1}, x^{k+1})\mid \cF_k \right] \\ &\leq F(w^0, x^0) - \sum_{t=0}^{k} \left[\frac{q'\gamma }{2\eta}\left\|\frac{\eta}{\gamma}\left(x^{t} - \overline{x}^{t+1}\right)\right\|^2 + \frac{q\tau}{2}\left\| \frac{1}{\tau}\left(w^{t} - \overline{w}^{t+1}\right)\right\|^2\right].
\end{align*}
\item \label{thm:smartconverges:item:convergencetostationarypoint} \textbf{Limit Points are Stationary.} Suppose that the sequence $\{(w^{k}, x^{k})\}_{k \in \NN}$ is almost surely bounded. Then $F(\overline{w}^{k+1}, \overline{x}^{k+1})$ converges almost surely to a random variable. Moreover, there exists a subset $\widetilde{\Omega} \subseteq  \Omega$ such that $P(\widetilde{\Omega}) = 1$ and for all $\omega \in \widetilde{\Omega}$, every limit point of $\{(\overline{w}^k(\omega), \overline{x}^k(\omega))\}_{k \in \NN}$ is  a stationary point of $F$.
\item \label{thm:smartconverges:item:sublinearrate} \textbf{Convergence Rate.} Fix $T \in \NN$. Sample $t_0$ uniformly at random from $t_0 \in \{0, \ldots, T\}$. Then 
\begin{align*}
 &\frac{q'\gamma}{2\eta}\EE\left[\left\|\frac{\eta}{\gamma}\left(x^{t_0} - \overline{x}^{t_0+1}\right)\right\| ^2 \right] + \frac{q\tau}{2} \EE\left[\left\| \frac{1}{\tau}\left(w^{t_0} - \overline{w}^{t_0+1}\right)\right\|^2\right]\leq \frac{F(w^0, x^0) - F(w^\ast, x^\ast)}{T}.\end{align*}
%(This result holds regardless of whether $\{(w^{k}, x^{k})\}_{k \in \NN}$ is bounded.)
\end{enumerate}
\end{theorem}

With proper choices of $b$, \textbf{we actually achieve an $\varepsilon$-accuracy solution with fewer gradient and function evaluations than the proximal gradient method or PALM~\citep{bolte2014proximal}}, which require $O(n/\varepsilon)$ gradient evaluations and $O(n/\varepsilon)$ function evaluations. 

The first corollary, whose proof is given Appendix~\ref{app:sublinear}, applies to a variant of the ProxSAGA algorithm: 

\begin{corollary}[Convergence Rate of SAGA Variant of SMART]\label{cor:sublinear}
Suppose that $D_k \equiv I_k$, $q' = 1 - 1/n$
\begin{align*}
\gamma = \frac{1}{4L\sqrt{\frac{1}{n}\sum_{i=1}^n   \frac{(1-1/n)(1+\epsilon_0) B_i^2}{2b\left(1 - \sqrt{(1-1/n)^{b+1}}\right)^2}} + \frac{L}{n}\sum_{i=1}^n B_i}, &&\text{and} && \tau = \frac{(n-1)\gamma}{\eta}.
\end{align*}
 Then SMART achieves an $\varepsilon > 0$ accurate solution with, on average, $O(n + n/(b^{3/2}\varepsilon) + n/(b^{1/2}\varepsilon))$ gradient evaluations,  $O(n/(b^{3/2}\varepsilon))$ evaluations of $\prox_{\gamma r_2}$, $O(n/(b^{3/2}\varepsilon))$ function evaluations, and $O(1/(b^{3/2}\varepsilon))$ evaluations of $\prox_{\tau r_1}$. In particular, when $b = n^{2/3}$, SMART achieves an $\varepsilon > 0$ accurate solution with, on average, $O(n + n^{2/3}/\varepsilon)$ gradient evaluations,  $O(1/\varepsilon)$ $\prox_{\gamma r_2}$ evaluations, $O(1/\varepsilon)$ function evaluations, and $O(1/(n\varepsilon))$ $\prox_{\gamma r_1}$ evaluations.\end{corollary}

The second corollary, whose proof is given Appendix~\ref{app:sublinear2}, applies to a variant of the ProxSVRG algorithm: 

\begin{corollary}[Convergence Rate of SVRG Variant of SMART]\label{cor:sublinear2}
Suppose that $D_k \equiv \emptyset$, $q' = (1 - 1/n)^b$,
\begin{align*}
\gamma = \frac{1}{4L\sqrt{\frac{1}{n}\sum_{i=1}^n   \frac{(1-1/n)^b(1+\epsilon_0 ) B_i^2}{2b\left(1 - \sqrt{(1-1/n)^{b}}\right)^2}} + \frac{L}{n}\sum_{i=1}^n B_i}, &&\text{and} && \tau = \frac{(1-(1-1/n)^b)(1-1/n)^b\gamma}{\eta}.
\end{align*}
 Then SMART achieves an $\varepsilon > 0$ accurate solution with, on average, $O(n + n/(b^{1/2}\varepsilon))$ gradient evaluations,  $O(n/(b^{3/2}\varepsilon))$ evaluations of $\prox_{\gamma r_2}$, $O(n/(b^{1/2}\varepsilon))$ function evaluations, and $O(1/(b^{1/2}\varepsilon))$ evaluations of $\prox_{\tau r_2}$. In particular, when $b = n^{2/3}$, SMART achieves an $\varepsilon > 0$ accurate solution with, on average, $O(n + n^{2/3}/\varepsilon)$ gradient evaluations,  $O(1/\varepsilon)$ $\prox_{\gamma r_2}$ evaluations, $O(n^{2/3}/\varepsilon)$ function evaluations, and $O(1/(n^{1/3}\varepsilon))$ $\prox_{\tau r_1}$ evaluations.\end{corollary}

\subsection{Linear Rates}\label{sec:linear}
Assuming that an error bound holds for all points $(w, x) \in \dom(r_1) \times \dom(r_2)$, a potentially bounded set, we can prove stronger convergence rates. 

\paragraph{The Global Error Bound.} In our analysis, we use a modified globalization of the error bound found in~\cite{drusvyatskiy2016error}. We assume that there exists $(w^\ast, x^\ast) \in \dom(r_1) \times \dom(r_2)$ such that for all $(w, x) \in \dom(r_1) \times \dom(r_2)$, we have
\begin{align*}
\mu \left[F(w, x) - F(w^\ast, x^\ast)\right] \leq &\left\|\frac{\eta}{\gamma}\left(x - \prox_{(\gamma/\eta) r_1}\left(x - \frac{\gamma}{\eta}\sum_{i=1}^n w_i\nabla f_i(x)\right)\right)\right\|^2 \\
+& \left\|\frac{1}{\tau}\left(w - \prox_{\tau r_2}(x - \tau (f_1(x), \ldots, f_n(x)))\right)\right\|^2 \numberthis\label{eq:error_bound} 
\end{align*}
\cite{drusvyatskiy2016error} use a localized version of~\eqref{eq:error_bound} to prove linear convergence of a proximal algorithm for minimizing convex composite objectives. Our error bound differs from their error bound in two ways: (1) their bound is only assumed to hold locally around critical points of $F$; and (2) their right hand side is $\mu \left[\|x - x^\ast\|^2 + \|w - w^\ast\|^2\right]$, rather than $\mu \left[F(w, x) - F(w^\ast, x^\ast)\right]$. We use this simplified error bound to keep the presentation short, but in future work, we may study the behavior of SMART assuming the localized bound in~\cite{drusvyatskiy2016error}.\footnote{Equation~\eqref{eq:error_bound} is also quite similar to the \textit{Kurdyka-{\L}ojasiewicz} (KL) inequality with exponent $\frac{1}{2}$~\citep{KLinequality,KLequality2}, which replaces the left hand side of~\eqref{eq:error_bound} by $\text{dist}(0, \partial_L F(w, x))^2$. Its straightforward to prove linear convergence of SMART under this globalized KL error bound, but we omit it to keep the presentation short.} 

As in the sublinear case, we define a constant $\eta$ (which depends on a user defined constant $\epsilon_0 \in (0, 1)$):
\begin{align}\label{eq:eta_definition2}
\eta &= 2 + 4\gamma\left[\sqrt{\frac{1}{n}\sum_{i=1}^n   \frac{q' (1+\epsilon_0)(B_iL)^2}{2b \sqrt{q'(1-\rho_{i})}\left(1 - (q'(1-\rho_{i}))^{1/4}\right)^2}} + \frac{L}{n}\sum_{i=1}^n B_i\right].%\\
\end{align}
The ratio $\gamma/\eta$ controls the linear convergence rate of SMART.

\begin{theorem}[Convergence Rate of SMART Assuming a Global Error Bound]\label{thm:convergencerate}
Assume the notation of Theorem~\ref{thm:smartconverges}. Let $\epsilon_0  \in (0, 1)$, let $\eta$ be defined as in~\eqref{eq:eta_definition2}, and let 
$$ 
\gamma = \frac{1}{4L\sqrt{\frac{1}{n}\sum_{i=1}^n \frac{q'(1+\epsilon_0) B_i^2}{2b\sqrt{q'(1-\rho_{i})}\left(1 - (q'(1-\rho_{i}))^{1/4}\right)^2}} + \frac{L}{n}\sum_{i=1}^n B_i}.
$$
Define 
$
\delta := \max_i\left\{1 - \mu\min\left\{\frac{q'\gamma}{2\eta}, \frac{q\tau}{2} \right\}, \sqrt{q'(1-\rho_i)}\right\}  \in (0, 1).
$
Then provided that the error bound~\eqref{eq:error_bound} holds, then we have
$$
\left(\forall k \in \NN\right) \qquad \EE\left[ F(w^{k}, x^{k}) - F(w^\ast, x^\ast) \right] \leq \delta^k\left[F(w^0, x^0) - F(w^\ast, x^\ast)\right].
$$
\end{theorem}

By assuming an error bound similar to~\eqref{eq:error_bound} and employing a restart strategy,~\cite{reddi2016fast} developed a linearly converging variant of ProxSAGA and ProxSVRG.
In this strategy, the authors ran ProxSAGA or ProxSVRG for $\ceil{30\kappa}$ iterations, where $\kappa$ is akin to the inverse condition number 
$
\kappa = L/\mu,
$ before restarting the algorithm. Every time that ProxSAGA or ProxSVRG is restarted, a full gradient must be computed. {\bf In contrast, SMART never needs to be restarted: it simply adapts to the regularity of the problem at hand.}

Frequent restarts of ProxSAGA and ProxSVRG lead to worse complexity. In both of the corollaries below, we show SMART needs $O(n + n^{2/3}\kappa\log(1/\varepsilon))$ gradients to reach accuracy $\varepsilon$. In contrast, ProxSAGA/SVRG need $O((n + n^{2/3}\kappa)\log(1/\varepsilon))$ gradients to reach accuracy $\varepsilon$.

The first corollary, whose proof is given Appendix~\ref{app:linear_saga}, applies to a variant of the ProxSAGA algorithm:

\begin{corollary}[Linear Convergence Rate of SAGA Variant of SMART]\label{cor:linear_saga}
Suppose that $D_k \equiv I_k$, $q' = 1 - 1/n$, that $\gamma$ is chosen as in Theorem~\ref{thm:convergencerate}, and $\tau = \frac{(n-1)\gamma}{\eta}$.
 Then SMART achieves an $\varepsilon > 0$ accurate solution with, on average, $O(n +\kappa (n/b^{3/2}+ n/b^{1/2})\log(1/\varepsilon))$ gradient evaluations,  $O((\kappa n/b^{3/2})\log(1/\varepsilon))$ evaluations of $\prox_{\gamma r_2}$, $O((\kappa n/b^{3/2})\log(1/\varepsilon))$ function evaluations, and $O((\kappa/b^{3/2})\log(1/\varepsilon))$ evaluations of $\prox_{\tau r_1}$. In particular, when $b = n^{2/3}$, SMART achieves an $\varepsilon > 0$ accurate solution with, on average, $O(n + \kappa n^{2/3}\log(1/\varepsilon))$ gradient evaluations, $O(\kappa\log(1/\varepsilon))$ $\prox_{\gamma r_2}$ evaluations, $O(\kappa\log(1/\varepsilon))$ function evaluations, and $O((\kappa/n) \log(1/\varepsilon))$ $\prox_{\tau r_1}$ evaluations.
 \end{corollary}
 
 The second corollary, whose proof is a straightforward modification of the proof of Corollaries~\ref{cor:linear_saga} and~\ref{cor:sublinear2}, applies to a variant of the ProxSVRG algorithm: 

\begin{corollary}[Linear Convergence Rate of SVRG Variant of SMART]\label{cor:linear_SVRG}
Suppose that $D_k \equiv \emptyset$, $q' = (1 - 1/n)^b$, that $\gamma$ is chosen as in Theorem~\ref{thm:convergencerate}, and that $\tau = \frac{(1-(1-1/n)^b)(1-1/n)^b\gamma}{\eta}.$ Then SMART achieves an $\varepsilon > 0$ accurate solution with, on average, $O(n + (\kappa n/b^{1/2})\log(1/\varepsilon))$ gradient evaluations,  $O((\kappa n/b^{3/2})\log(1/\varepsilon))$ evaluations of $\prox_{\gamma r_2}$, $O((n/b^{1/2})\log(1/\varepsilon))$ function evaluations, and $O((\kappa/b^{1/2})\log(1/\varepsilon))$ evaluations of $\prox_{\tau r_2}$. In particular, when $b = n^{2/3}$, SMART achieves an $\varepsilon > 0$ accurate solution with, on average, $O(n + \kappa n^{2/3}\log(1/\varepsilon))$ gradient evaluations, $O(\kappa \log(1/\varepsilon))$ $\prox_{\gamma r_2}$-proximal operator evaluations, $O(\kappa n^{2/3}\log(1/\varepsilon))$ function evaluations, and $O((\kappa/n^{1/3})\log(1/\varepsilon))$ $\prox_{\tau r_1}$ evaluations.\end{corollary}

\section{Numerics}
\label{sec:numerics}

In this section we perform trimmed model fitting (i.e., we solve~\eqref{eq:trimmed} with a regularizer) on three models/datasets:
\begin{enumerate}
\item recognizing hand-written digits (0-9) with multinomial classification on the {\bf mnist} dataset~\citep{lecun1998gradient};
\item trimmed principal component analysis, using the {\bf US Judges} dataset provided in \verb{R{~\citep{team2013r};
\item robust homography estimation using interest point matching.
\end{enumerate}

The latter two applications are formulated using nonconvex constraints.  
Plots for figures~\ref{fig:Ratecomp} and~\ref{fig:t-pca} were generated with Matplotlib~\citep{Hunter:2007}.

\subsection{Multi-class classification}
\label{sec:multi-class}

The {\bf mnist} training dataset contains 60000 pictures of hand-written digits between 
0-9. We model automated digit recognition as a multi-class classification problem 
with $K=10$ classes. We briefly review multinomial logistic regression to align~\eqref{eq:trimmed} with our current formulation.
%%%%%%%%%%%%%%%%%%%%%%%%%%%%%%%%%%%
\paragraph{Formulation:}
We are given $n$ data pairs $(v_i, y_i)$, where $v_i \in \mathbb{R}^p$
are training features, and $b_i \in \mathbb{R}^K$ are `one-hot' training labels. 
If the $i$th example belongs to the $j$th class, then $y_i = e_j$, the $j$th standard unit vector.  

The decision variable is a matrix $X \in \mathbb{R}^{p \times K}$ and  
each column $x_j$ of $X$ defines a linear classifier. 
The soft-max loss is a standard objective for selecting the best fitting classifier out of a given set: 
%\[
$f_i(X) = \log(\sum_{j=1}^K \exp(\langle v_i,x_j\rangle)) - v_i^T X y_i.$
%\]
Define the log-sum-exp(LSE) function by $\LSE(z) = \log\left(\sum_j \exp(z_j)\right)$.
%\[
%f_i(X) =
%\log\left(\sum_{j=1}^K \exp(\langle v_i, x_j\rangle)\right) - \sum_{j=1}^k \langle e_j, y_i \rangle\langle v_i, x_j\rangle. 
%\]
\begin{figure}[h!]
\center
\begin{tabular}{cccccc}\\ 
\includegraphics[scale=0.1]{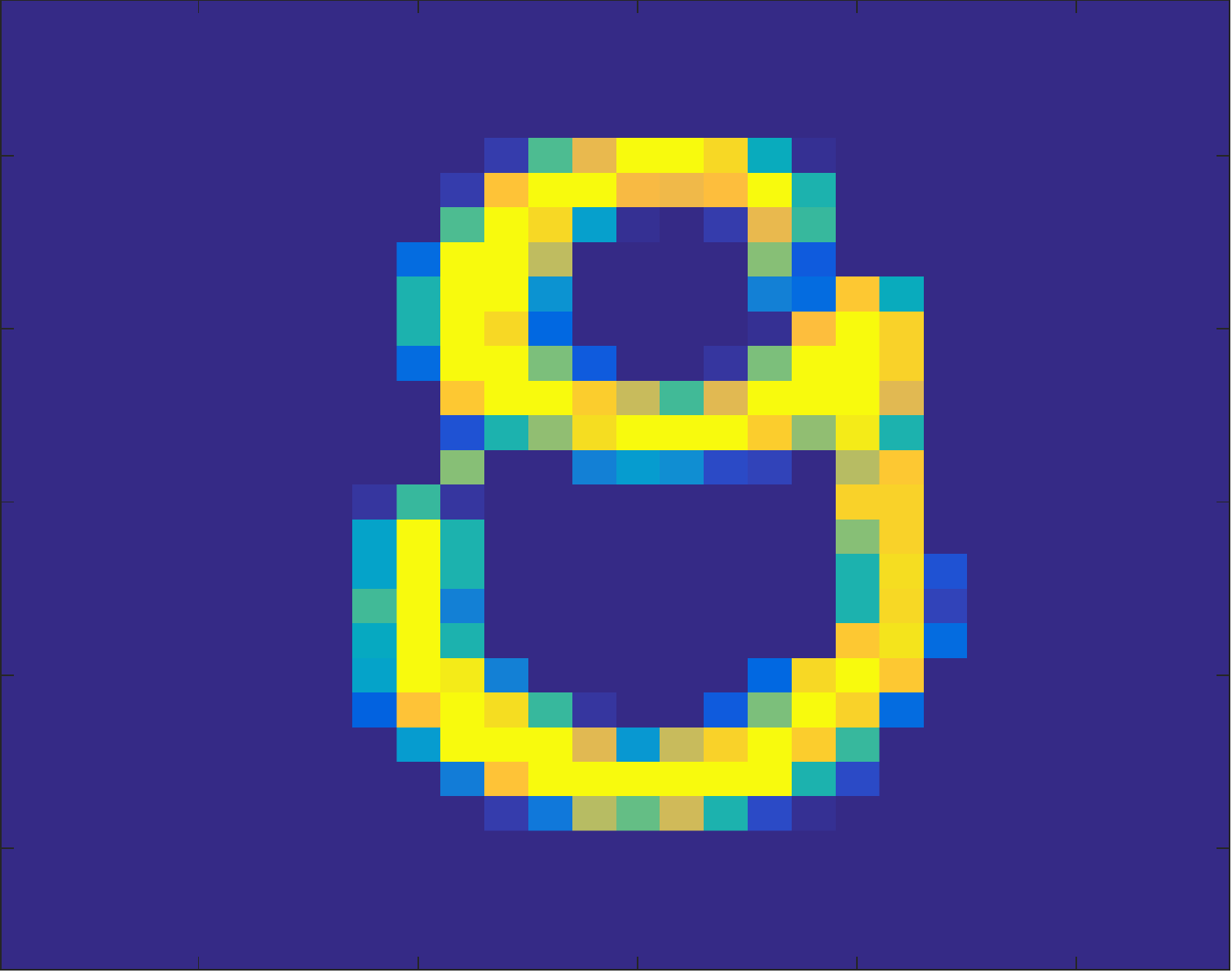}
&\includegraphics[scale=0.1]{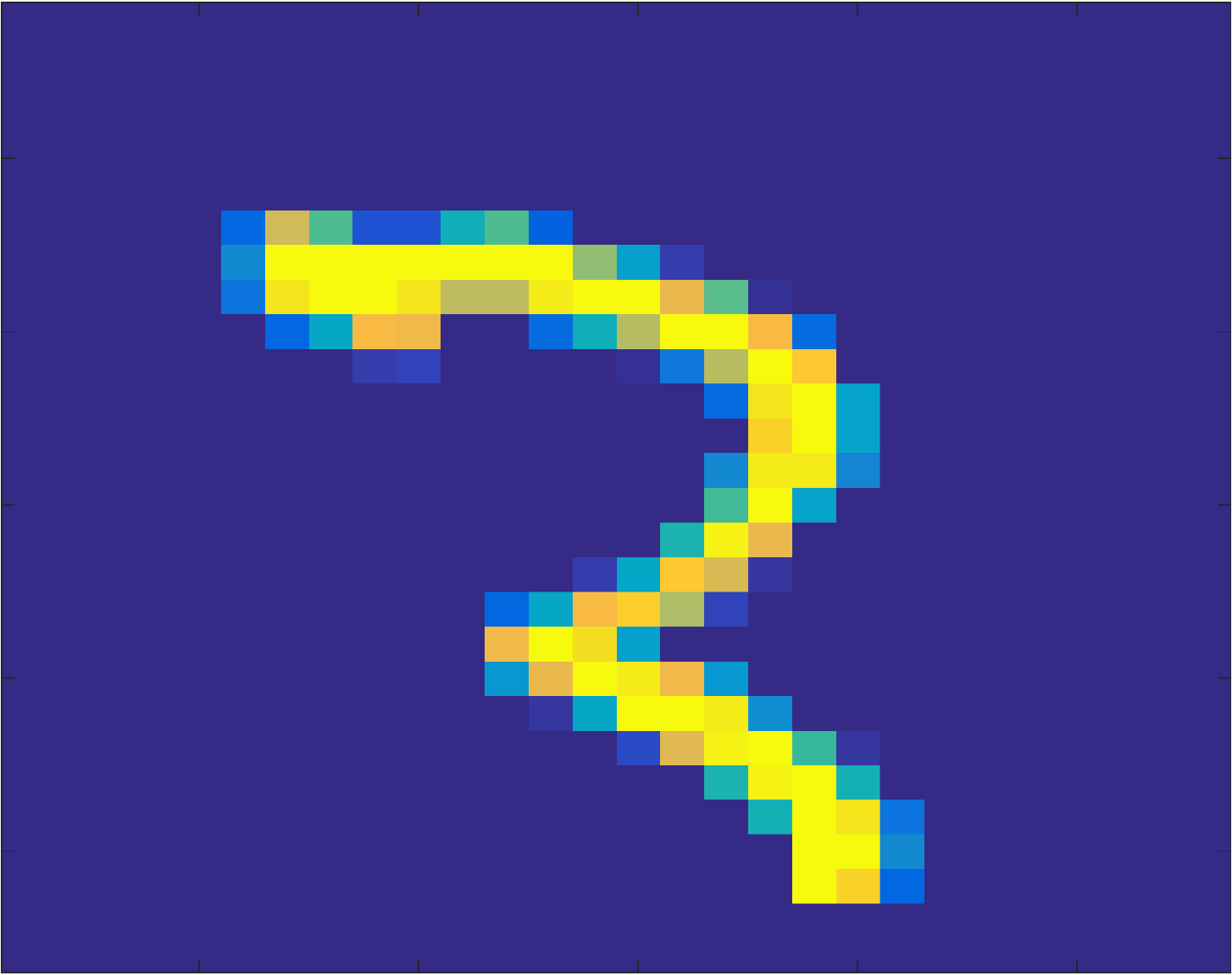}
&\includegraphics[scale=0.1]{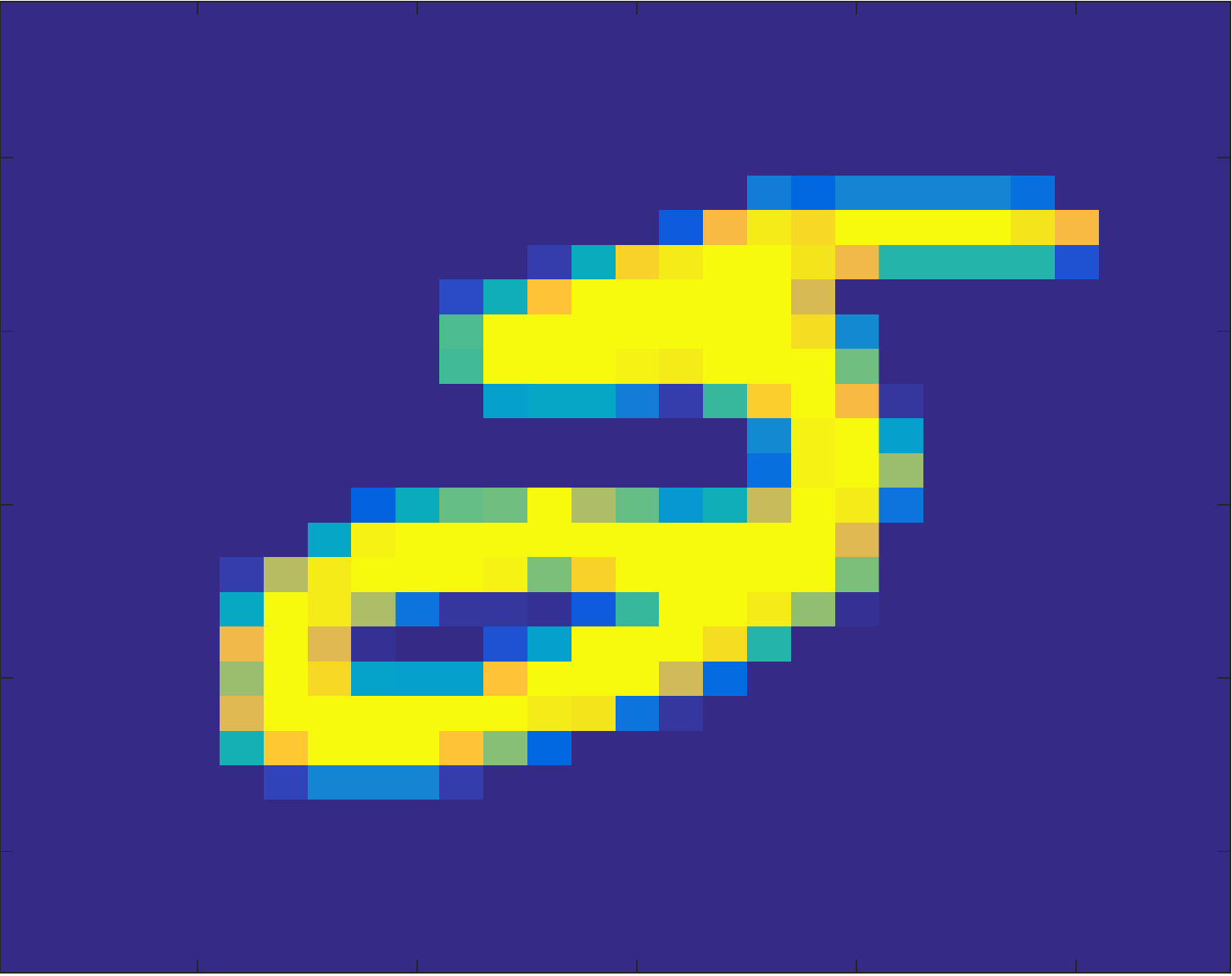}
&\includegraphics[scale=0.1]{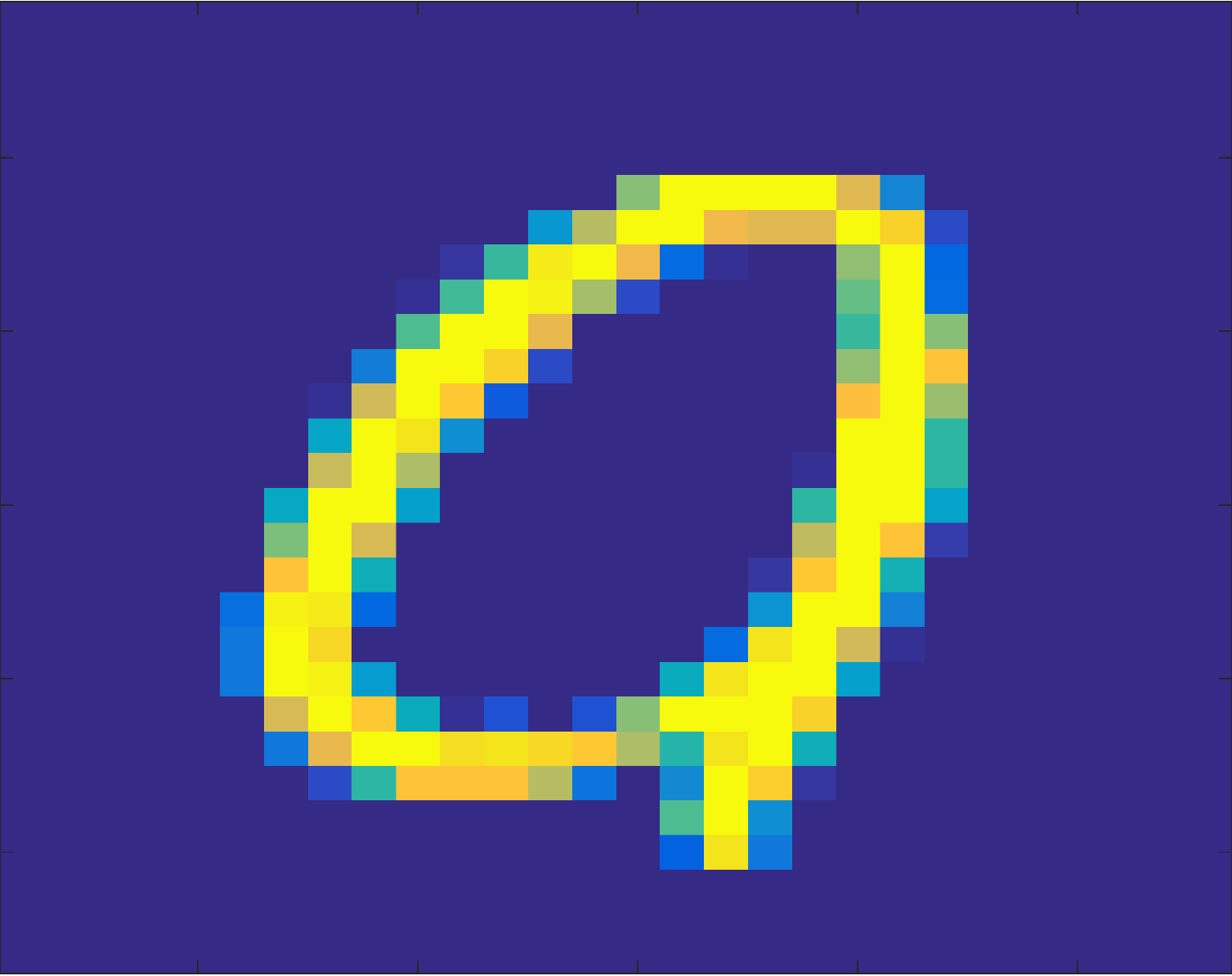}
&\includegraphics[scale=0.1]{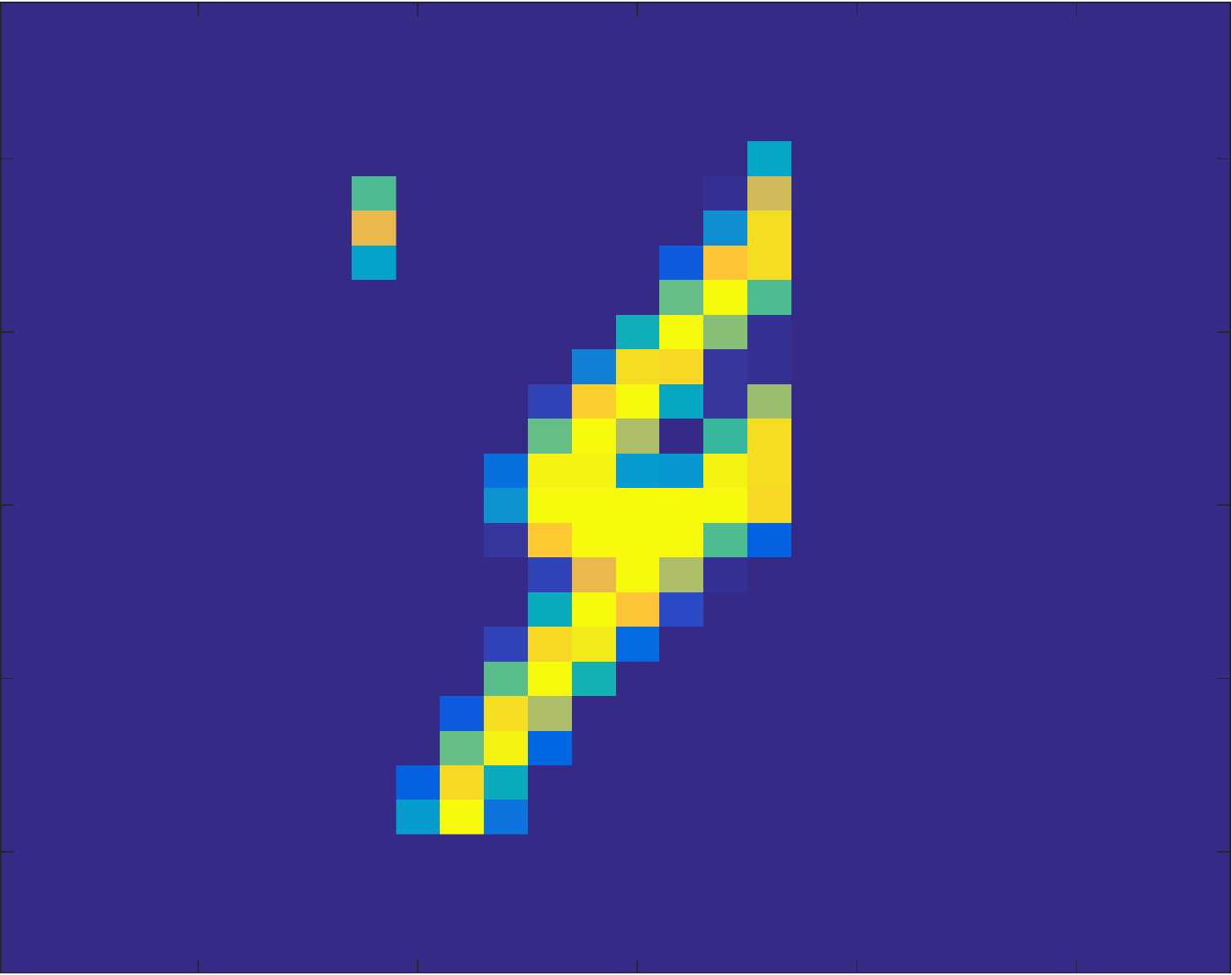}
&\includegraphics[scale=0.1]{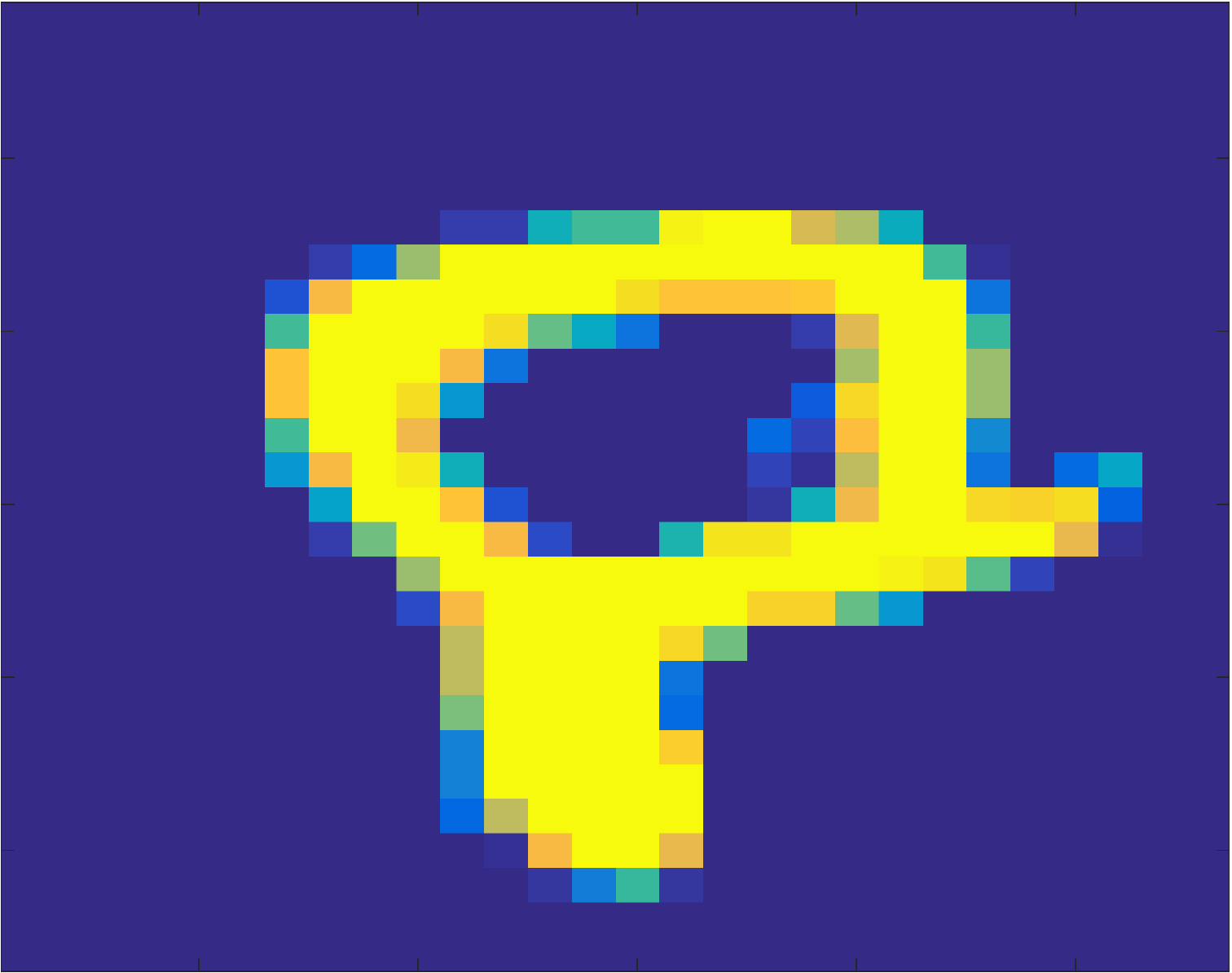}\\
? & ? & ? & ? & ? &?
\end{tabular}
\caption{\label{fig:mnistNat} Six of the outliers found in the original MNIST dataset using trimming. 
Can you guess what the labels are? (See text for answers.)} %(They are given in the text describing the experiment). }
\end{figure}
%
%The LSE function is smooth, with gradient %given by and hessian given by 
%\[
%\begin{aligned}
%\nabla_{x_j} f_i
%& = 
%\left(
%\frac{\exp(\langle v_i, x_j\rangle)}{\exp(\LSE(Xv_i))} -  y_{ij}\right) v_i.
%%\\
%%\nabla^2 \LSE(z) &= \frac{1}{\exp(\LSE(z))}\mbox{diag}(\exp(z)) - \frac{1}{\exp(\LSE(z))^2} \exp(z)\exp(z)^T, 
%\end{aligned}
%\]
%and so the Lipschitz constant of $\nabla \LSE$ is no greater than 2. 
%
The trimmed (regularized) multiclass problem is given by 
\begin{equation}
\label{eq:LSEtrim}
\min_{X} \min_{w\in\Delta^m} \frac{1}{n}\sum_{i=1}^n w_i\left(\LSE(Xv_i) - v_i^TXy_i\right) + R(X). 
\end{equation}
For simplicity, we use $R(X) = \frac{\lambda}{2n} \|X\|^2$, where $n$ is the number of examples. 
%
%%%%%%%%%%%%%%%%%%%%%%%%%%%%%%%%%%%
\paragraph{Experiments:} We use $\lambda = 0.01$ consistently for all experiments.
We first set $m = 0.998 n$, to find outliers in the actual {\bf mnist} dataset. 
Figure~\ref{fig:mnistNat} depicts these outliers. Visually, the labels are hard to decipher, but their assigned labels are, from left to right, $9, 3, 5, 9,4,8$.
Studying outliers, once they are detected, can give interesting insights into the learning example. 

Messily written digits plague {\bf mnist} training and test sets, so we should not expect that removing potential outliers from the training set  improves classification performance 
on the test set. However, when we maliciously contaminate the {\bf mnist} training set by shifting a large portion of the labels by 1 
(modulo 9), trimming accuracy degrades only slightly, while the standard approach fails dramatically.

We show the effects of malicious contamination in Table~\ref{mnistOut}. For the trimmed formulation, 
we always over-estimate the proportion of outliers by 10\%. 
Then, we evaluate the predictive accuracy of the trimmed and standard approaches on the test set. We also evaluate how well each method detects outliers.

For the standard approach, we fit the untrimmed LSE model and then label as outliers the data points which obtain the $n-h$ largest objective values. This approach is standard in regression. 
For the trimmed method, the outliers are determined by the zero-set of the $w$ vector. 

\begin{table}
\parbox{\linewidth}{
\centering
\begin{tabular}{ll|llll|l}
 \toprule
\textbf{Outliers}&\textbf{0\%}& \textbf{10\%}&\textbf{20\%}&\textbf{30\%} & \textbf{40\%}&\textbf{50\%}\\\cmidrule{1-7}
LSE-Accuracy &92.28&89.2& 85.3&78.8&65.4&44.9\\%\hline
LSE-Detection &---   &90.8&90.4 &82.4&71.8&61.0\\
LSE-False-Pos &---&11.5&14.9 &21.8&35.5&59.0\\\hline
SMART-Accuracy &91.2&90.7 &89.9&89.0&86.8&43.7\\%\hline
SMART-Detection &---&99.6 &99.1&98.2&96.8&61.7\\%\hline
SMART-False-Pos &---&11.4 &12.7&16.4&19.5&58.6\\%\hline
\bottomrule
\end{tabular}
\caption{\label{mnistOut} Accuracy, detection, and false positive rate for standard (LSE) approach and trimmed approach. The approximate number of outliers, required by SMART, is over-estimated by 10\%
to reflect a realistic application of the method. The first column shows that over-estimation of outliers by 10\% even in the nominal case carries only a $1\%$ cost in terms of predictive accuracy.}
}
\end{table}
The results are shown in Table~\ref{mnistOut}. While the trimmed formulation (solved with SMART)
degrades only slightly with between 10\%- 40\% systematic contamination, the standard approach degrades much more rapidly. 
Even with 40\% mislabeled data, 
SMART is able to identify more than 95\% of the outliers that we  maliciously injected.  

%
%The extreme case of systematic contamination is shown in the last column of Table~\ref{mnistOut}. 
When the proportion of systematic errors reaches 50\%, both methods degrade rapidly. This is not surprising: when 50\% of labeled data is both wrong and mutually consistent, 
we are just as likely to find the incorrect model. 
%However, if more than 50\% of the data are both wrong and mutually consistent, there is little one can do. 

\begin{figure}[t!]
    \begin{subfigure}[t]{0.5\textwidth}
        \centering
        \includegraphics[scale=0.3]{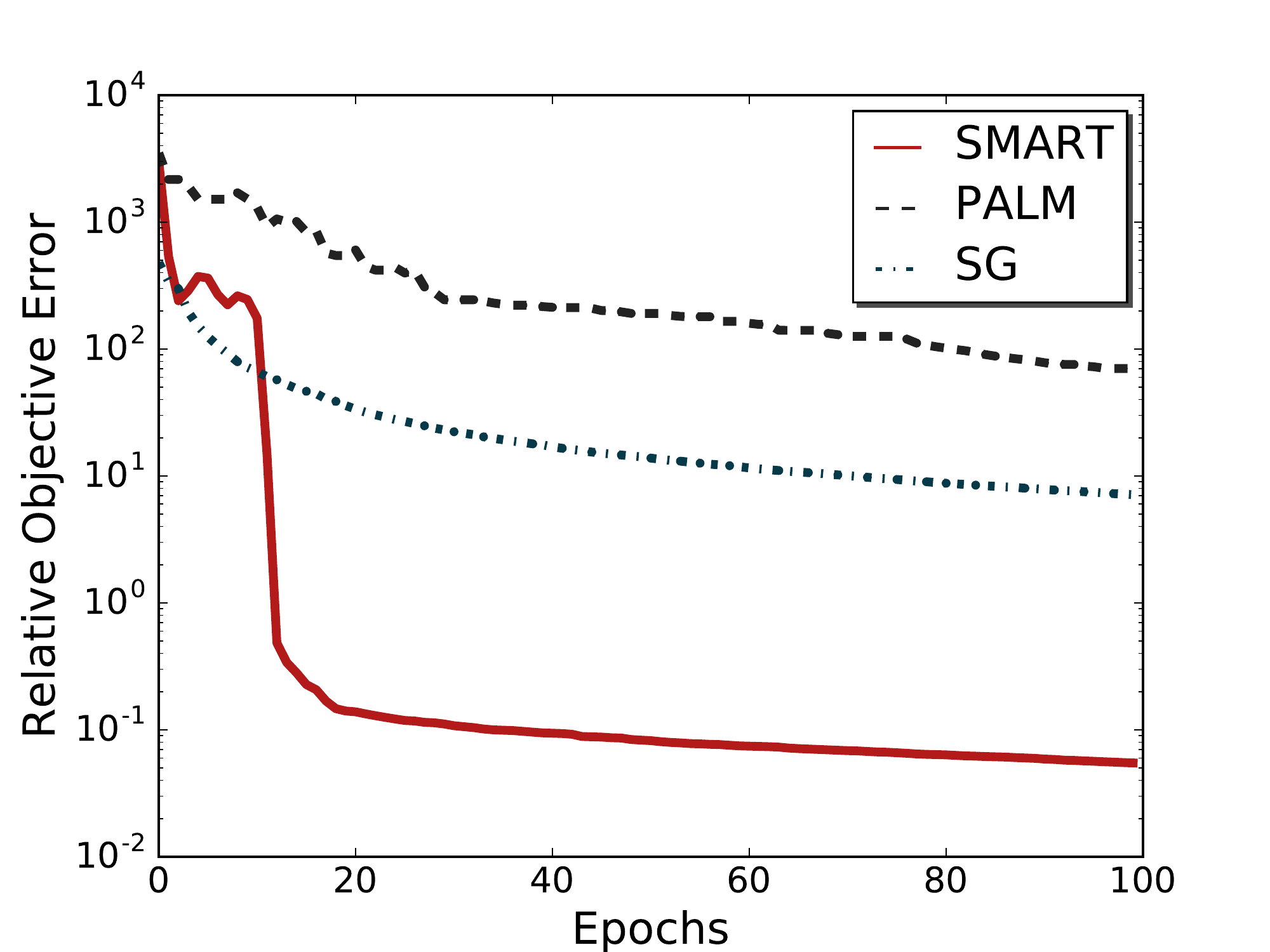} 
        \caption{10\% contamination}
    \end{subfigure}%
    ~ 
    \begin{subfigure}[t]{0.5\textwidth}
        \centering
        \includegraphics[scale=0.3]{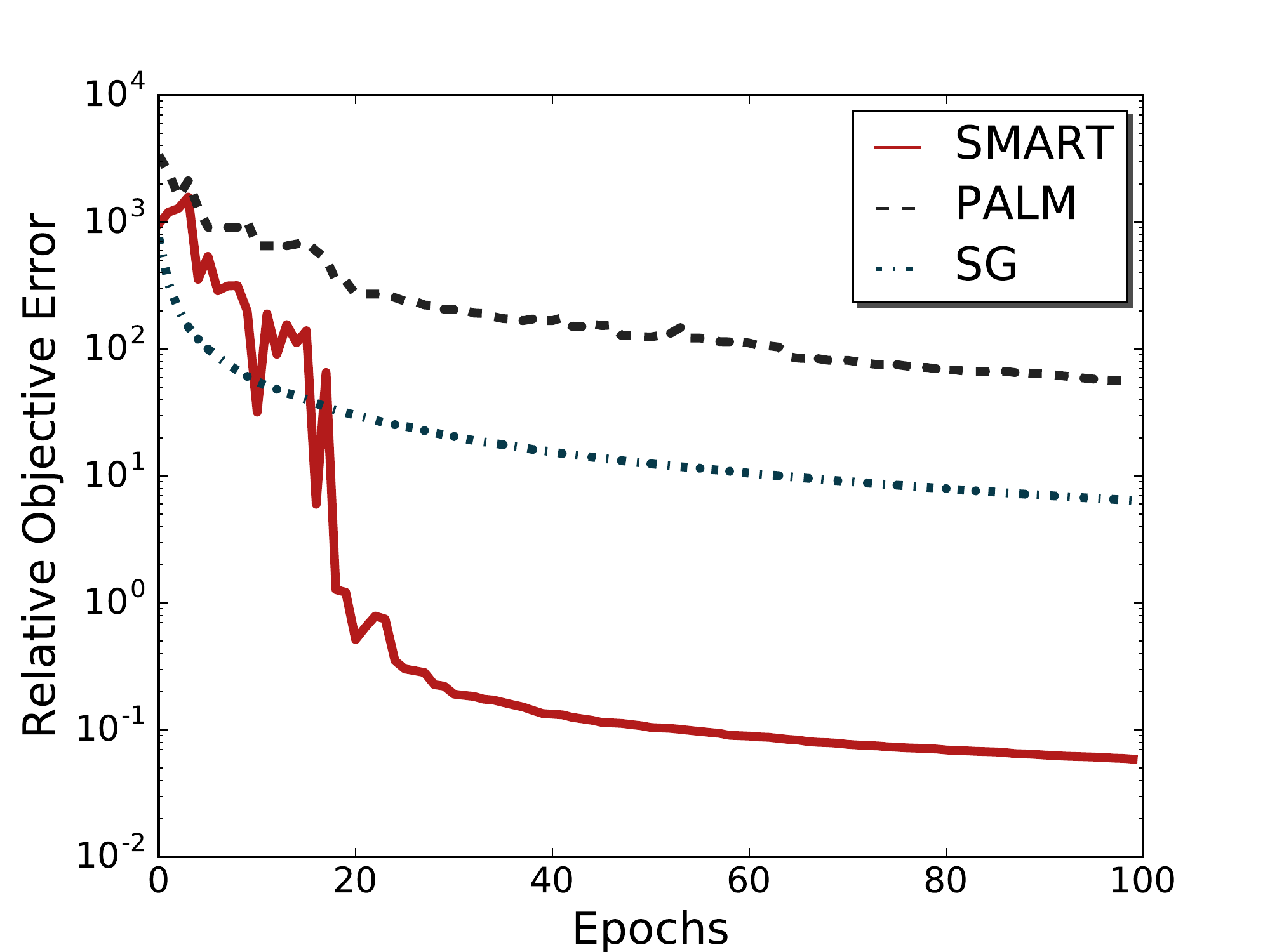} 
        \caption{20\% contamination}
    \end{subfigure}\\
        \begin{subfigure}[t]{0.5\textwidth}
        \centering
        \includegraphics[scale=0.3]{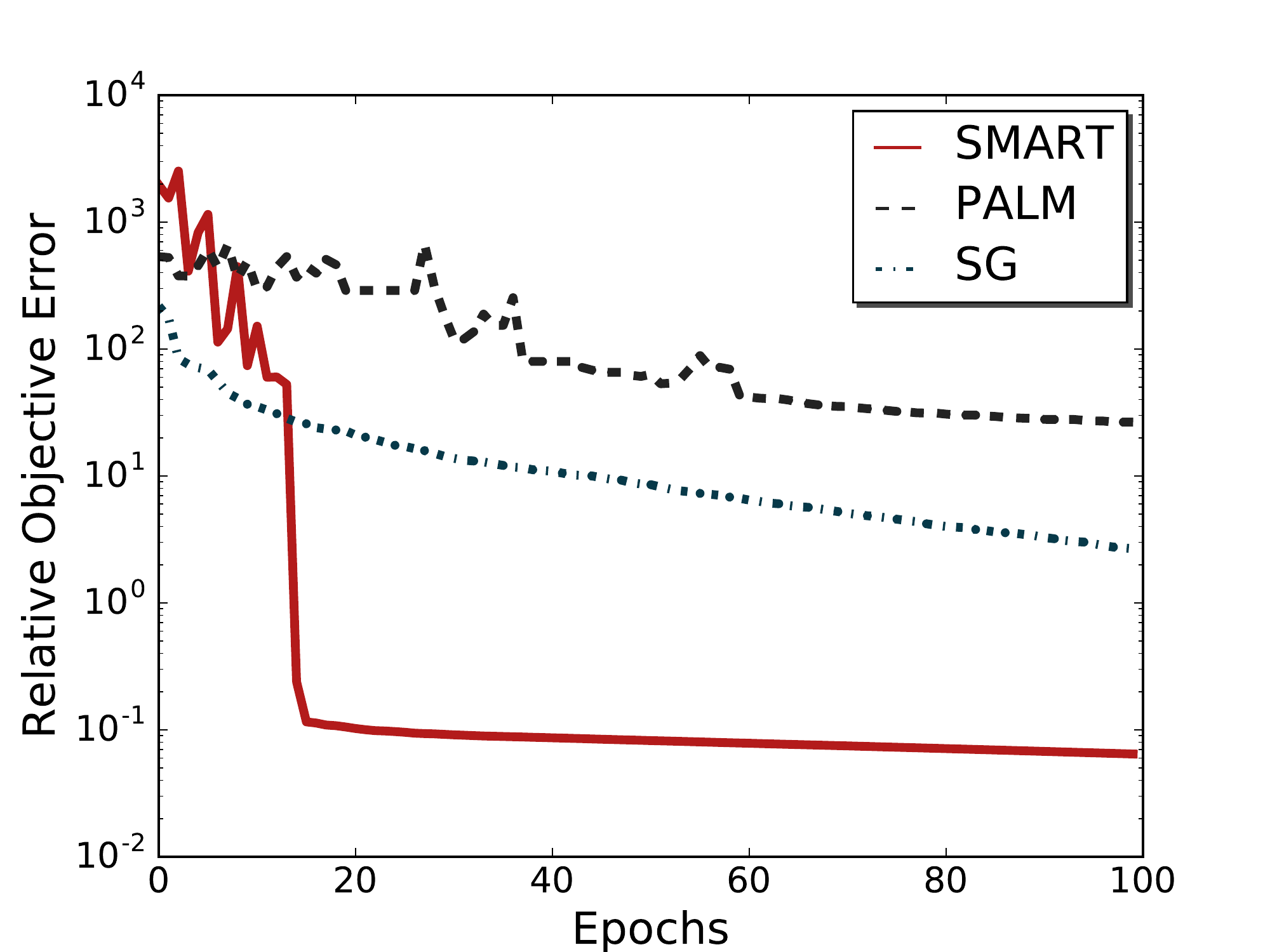} 
        \caption{30\% contamination}
        \end{subfigure}
        ~
        \begin{subfigure}[t]{0.5\textwidth}
        \centering
        \includegraphics[scale=0.3]{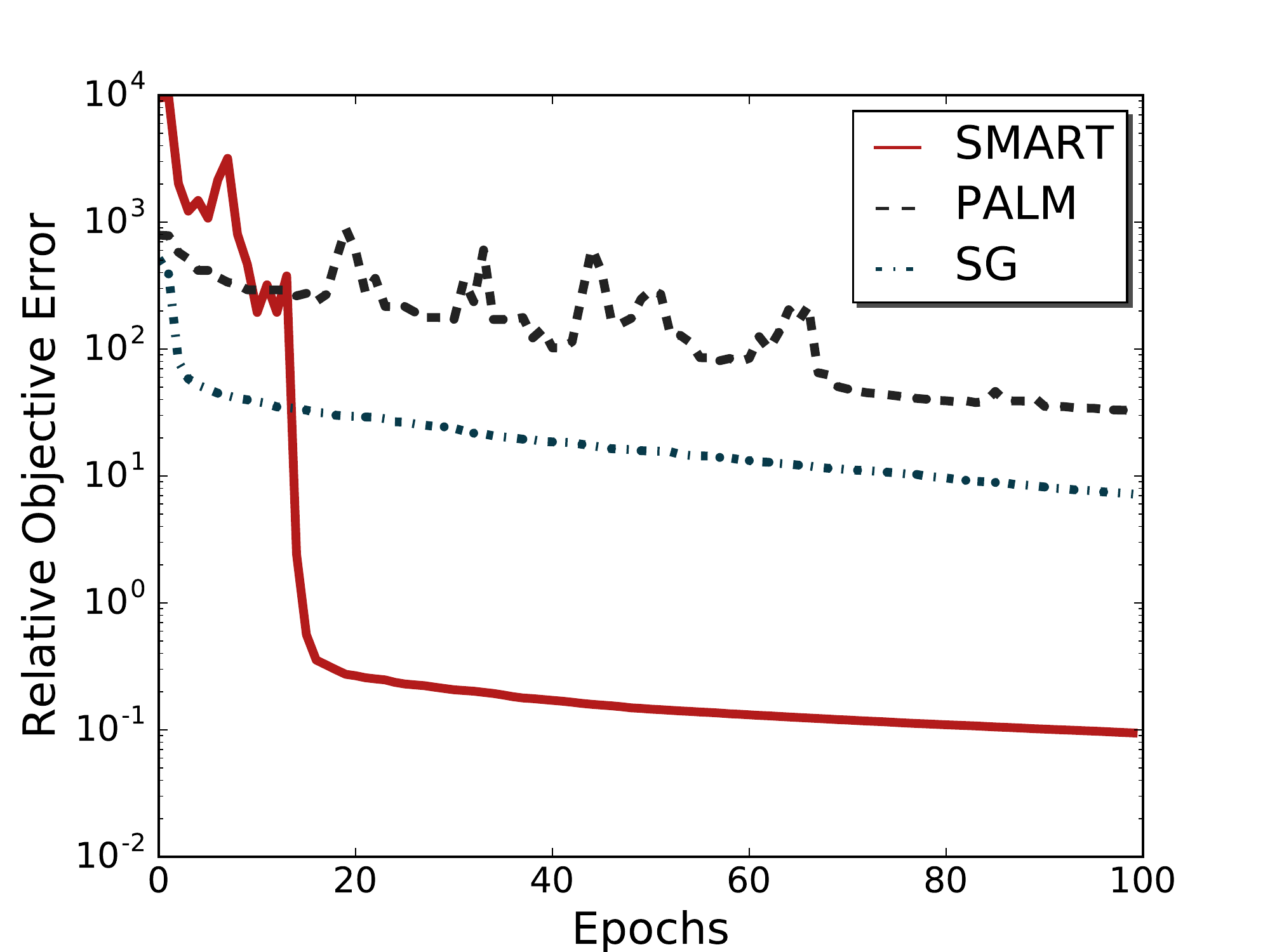} 
        \caption{40\% contamination}
    \end{subfigure}
    %\caption{30\% contamination}
%\center
%\includegraphics[scale=0.6]{RateComp} 
\caption{\label{fig:Ratecomp} 
Comparison of SMART (SVRG variant in Corollary~\ref{cor:sublinear2}), PALM~\citep{bolte2014proximal}, and a SG with minibatching (of size $n^{2/3}$) in terms of relative objective error $(F(w^k, x^k) - F(w^\ast, x^\ast))/F(w^\ast, x^\ast)$, which is not computed at every iteration, but only at the start of each epoch (i.e., after each full pass through all 60000 datapoints). In each of the four subplots, we maliciously contaminated a certain portion of the training labels as discussed in the text.}
%by adding to each the number 1 modulo 9. Then, we solve the trimmed formulation~\eqref{eq:LSEtrim}. In all of the experiements, we overestimate the number of contaminated points by 10\%. We show the classification performance of the SMART estimator in Table~\ref{mnistOut}.}
\end{figure}

%%%%%%%%%%%%%%%%%%%%%%%%%%%%%%%%%%%
\paragraph{Performance comparison with PALM and SG.} 
In Figure~\ref{fig:Ratecomp}, we compare SMART to PALM~\citep{bolte2014proximal} and SG. In all of our experiments, we manually found the best stepsizes $\gamma$ and $\tau$ for PALM, SMART, and SG. We chose SMART's batch size to be $b := \ceil{n^{2/3}}= 1533$. For a fair comparison, we ran SG with a minibatch of the same size. Because~\eqref{eq:LSEtrim} is nonsmooth and nonconvex, there is no method to determine the global minimizer $(w^\ast, x^\ast)$ of $F$. As a proxy for $F(w^\ast, x^\ast)$, we ran SMART multiple times, for many iterations, and chose the lowest achieved objective value. We found that although PALM and SG are competitive with SMART during the first few passes through the dataset, their performance quickly stagnates, possibly due to finding spurious stationary points.

%%%%%%%%%%%%%%%%%%%%%%%%%%%%%%%%%%%%%%%%%%%%%%%%

\subsection{Trimmed Principal Component Analysis}

For a given matrix $A \in \mathbb{R}^{m \times n}$, we can analyze its principal linear components by finding, in the least squares sense, the best rank $k$ approximation to $A$. The principal components of $A$ are found through the singular value decomposition
\begin{equation}
\label{eq:SVD}
A = UDV^T,
\end{equation}
where $U \in \mathbb{R}^{m\times k}$ and $V\in\mathbb{R}^{m\times k}$ are orthogonal matrices, while 
$D \in \mathbb{R}^{k\times k}$ is diagonal with non-negative entries. The columns of the matrix
$X = UD$ are the \textit{principal components} of $A$ and $V$ is their corresponding {\it loadings}.  This process of finding $U, V, D$ and $X$ is called \textit{Principal Component Analysis} (PCA).

\paragraph{Formulation.}
It is well known that the matrix $U$ in PCA minimizes
\begin{align}\label{eq:pca}
\min_{U\in \mathcal{O}^{m\times k}} \frac{1}{2}\|(I-UU^T) A\|^2,
\end{align}
where $\mathcal{O}^{m \times k}$ is the set of $m \times k$ matrices with orthonormal columns. Trimmed-PCA seeks such a $U$ while simultaneously removing the influence of potentially contaminated columns $a_i$ of $A$:
\begin{align*}
\min_{w \in \Delta^h, U\in \mathcal{O}^{m\times k}}
&\frac{1}{n} \sum_{i=1}^n  \frac{w_i}{2}\left\|(I-UU^T)a_i\right\|^2. \numberthis \label{eq:tpca}%%\\&=\frac{1}{n} \sum_{i=1}^n \frac{w_i}{2} \left(\|a_i\|^2 - \|U^Ta_i\|^2\right).  
\end{align*}
Note that $U \in \cO^{m \times k}$, implies that 
$
\left\|(I-UU^T)a_i\right\|^2  = \|a_i\|^2 - \|U^Ta_i\|^2.
$
Thus, the PCA loss function is the sum of concave functions (each with a Lipschitz continuous derivative), while the regularizer $r_2$ is the indicator function of the orthogonal manifold $\cO^{m\times k}$. When combined with trimming, PCA is highly nonconvex. Nevertheless, by Theorem~\ref{thm:smartconverges}, SMART will converge when applied to this problem because the iterates $U^k$ lie in the bounded set $\cO^{m\times k}$.

Although it may seem that computing $\prox_{\gamma r_2} = P_{\cO^{m\times k }}$ dominates the cost of SMART on the trimmed-PCA problem, in reality the condition $k \approx b$ ensures that the costs of gradient and projection steps are balanced. Indeed, each batch gradient with $b= n^{2/3}$ samples requires $O(kmn^{2/3})$ arithmetic operations, while each $U$-projection requires only $O(mk^2)$ operations.

\paragraph{Experiments}

\begin{figure}[t!]
    \begin{subfigure}[t]{0.5\textwidth}
        \centering
        \includegraphics[scale=0.3]{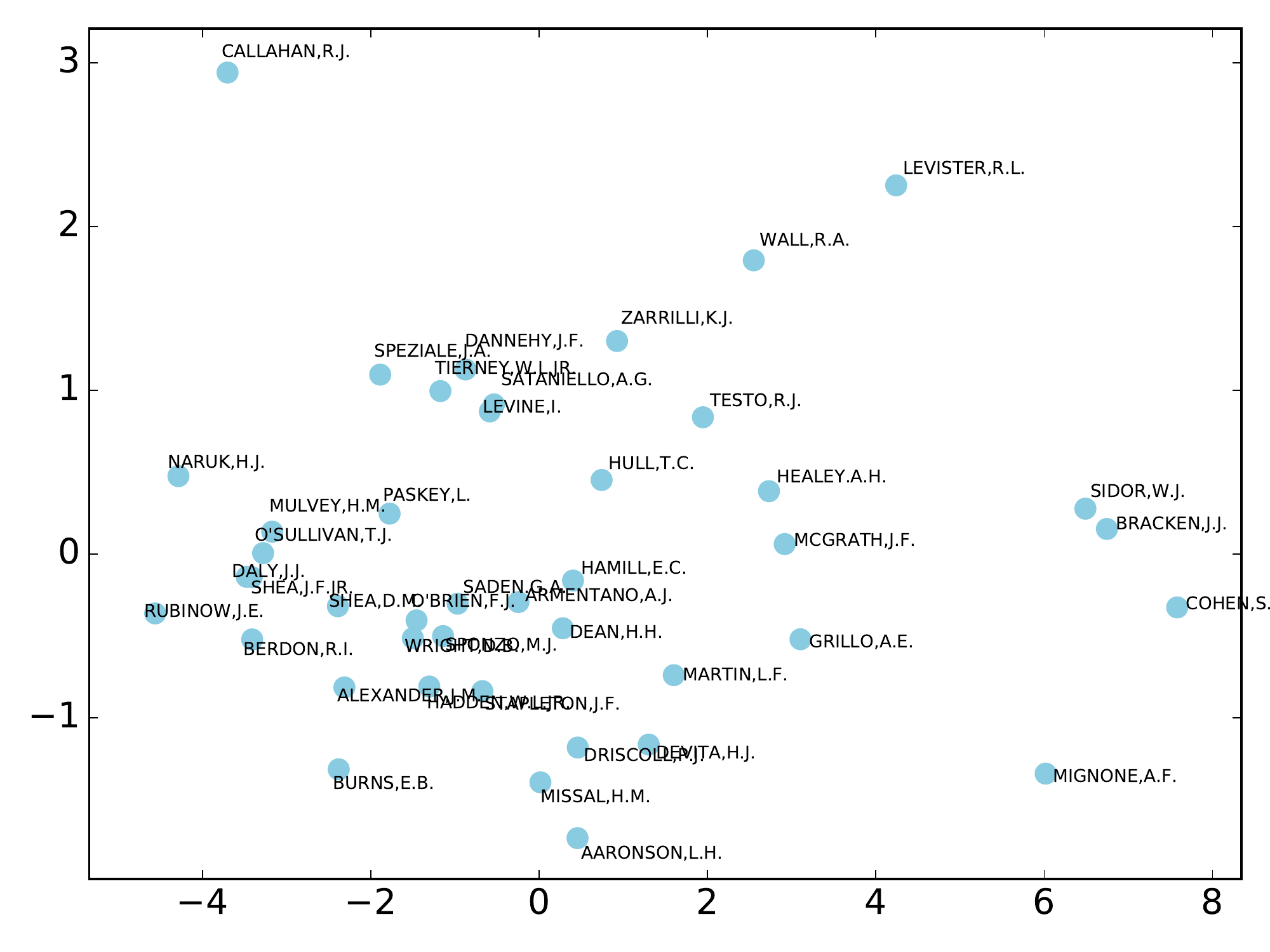} 
        \caption{PCA on Full Dataset}\label{subfig:PCA}
    \end{subfigure}%
    ~ 
    \begin{subfigure}[t]{0.5\textwidth}
        \centering
        \includegraphics[scale=0.3]{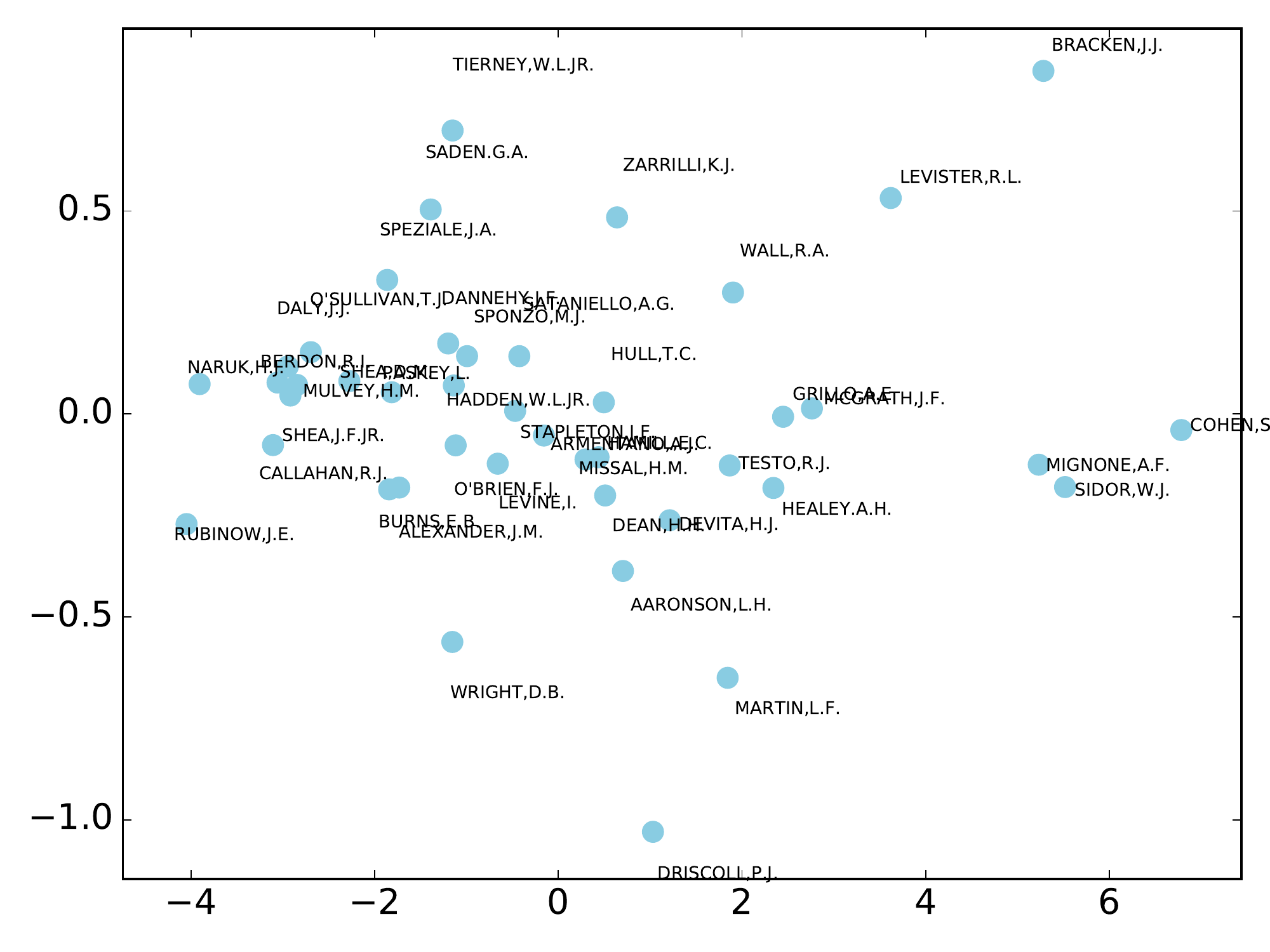} 
        \caption{PCA After Uninformative Categories Removed}\label{subfig:PCA-removed}
    \end{subfigure}\\
        \begin{subfigure}[t]{0.5\textwidth}
        \centering
        \includegraphics[scale=0.3]{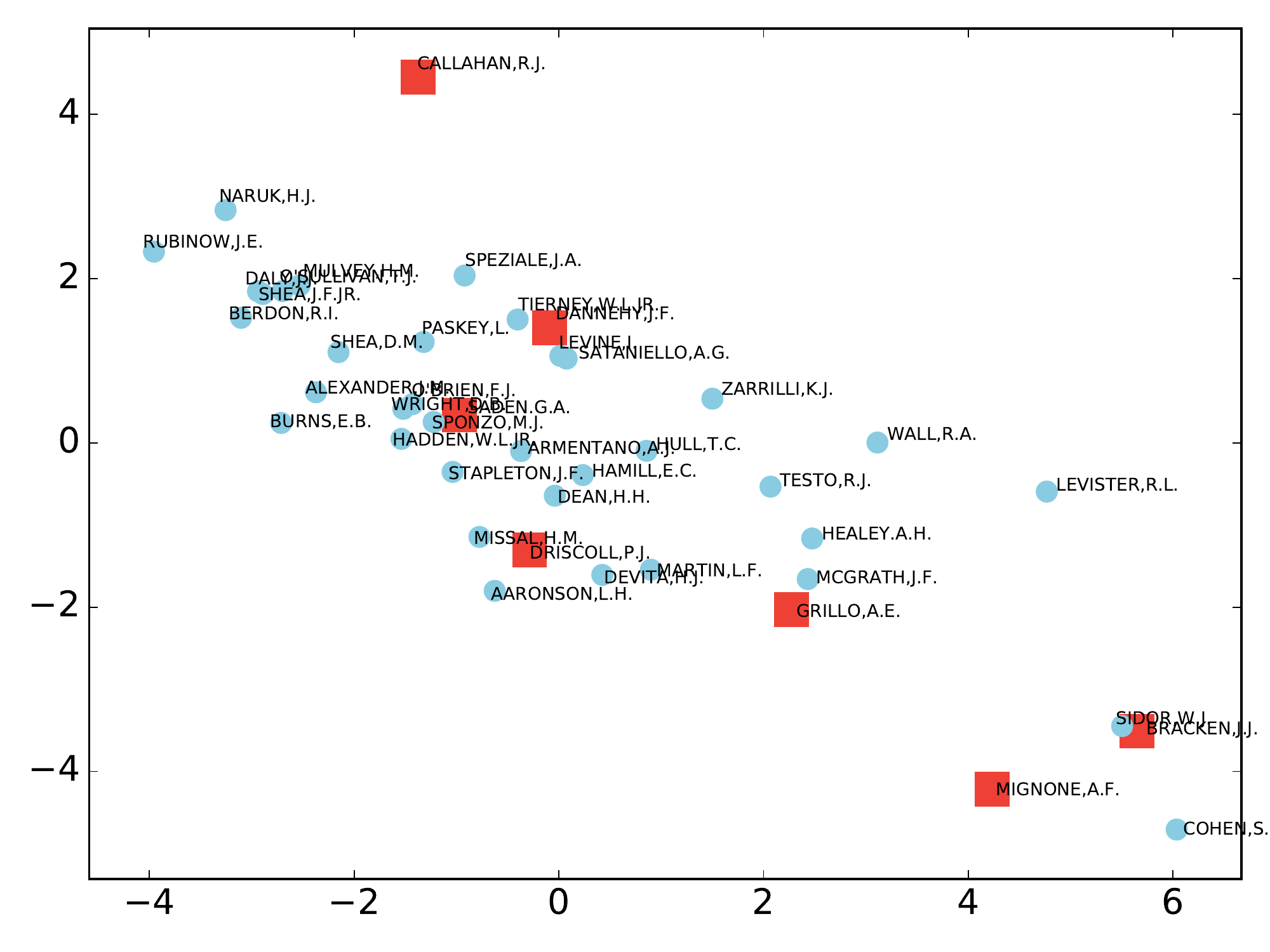} 
        \caption{20\% Trimming on Full Dataset}\label{subfig:T-PCA-Full}
        \end{subfigure}
        ~
        \begin{subfigure}[t]{0.5\textwidth}
        \centering
        \includegraphics[scale=0.3]{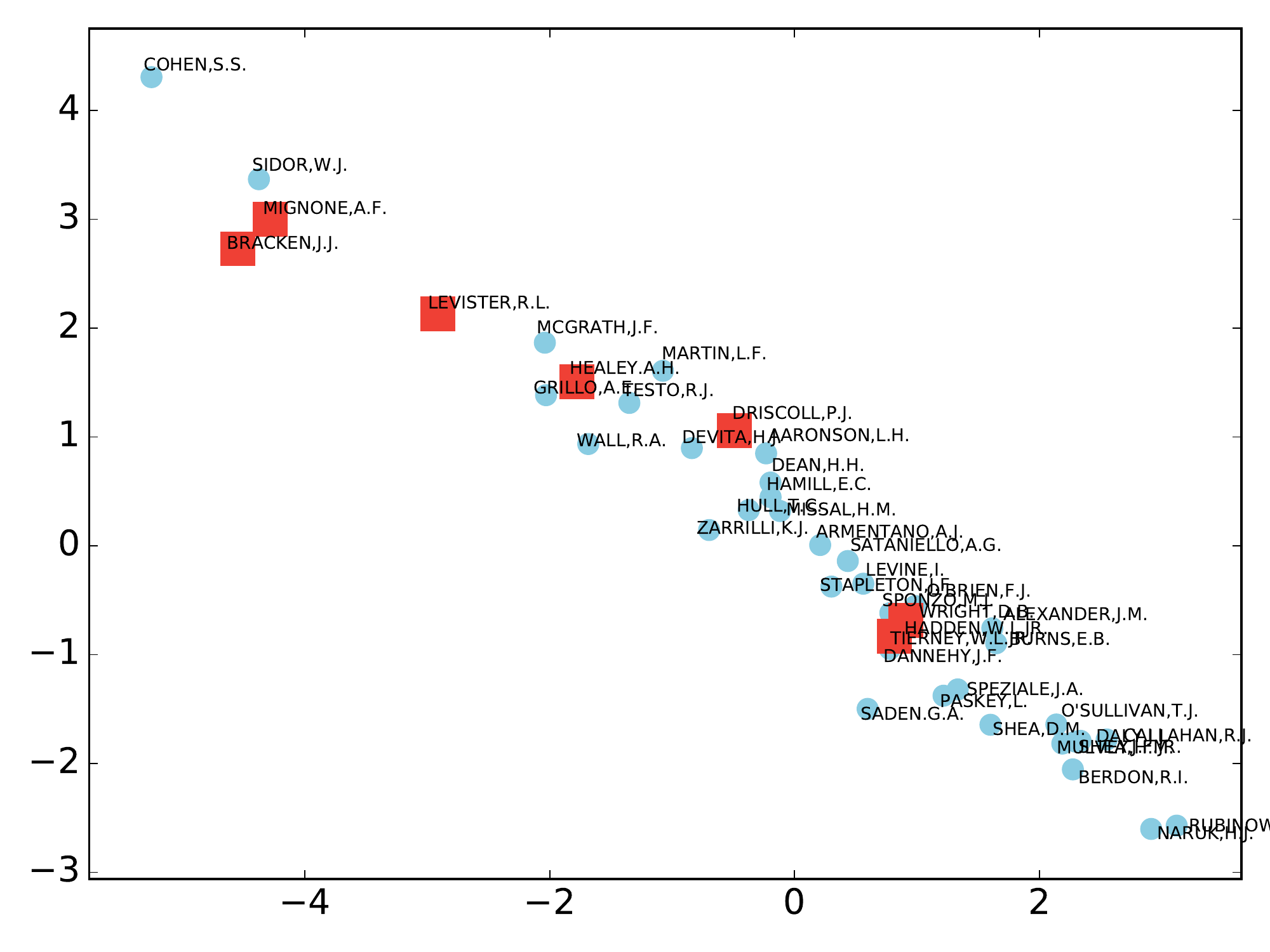} 
        \caption{20\% Trimming After Uninformative Categories Removed}\label{subfig:T-PCA-removed}
    \end{subfigure}
    %\caption{30\% contamination}
%\center
%\includegraphics[scale=0.6]{RateComp} 
\caption{\label{fig:t-pca} 
PCA and trimmed PCA on the {\bf US Judges} dataset. The left column depicts PCA and trimmed PCA on the full data matrix $A \in \RR^{12\times 43}$,
%(i.e., the datasets $U_{A, \text{PCA}}^T A \in \RR^{2 \times 43}$ and $U_{A, \text{T-PCA}}^TA \in \RR^{2\times 43}$, where $U_{A, \text{PCA}}$ and $U_{A, \text{T-PCA}}$ are found in~\eqref{eq:pca} and~\eqref{eq:tpca}), 
while the right column depicts PCA and trimmed PCA on the reduced data matrix $B \in \RR^{8\times 43}$;
%(i.e., the datasets $U_{B, \text{PCA}}^TB \in \RR^{2 \times 43}$ and $U_{B, \text{T-PCA}}^TB \in \RR^{2\times 43}$, where $U_{B, \text{PCA}}$ and $U_{B, \text{T-PCA}}$ are found in~\eqref{eq:pca} and~\eqref{eq:tpca}); 
see the text for a description of these matrices.  
}
\end{figure}

We used the {\bf US judges} datset to test trimmed-PCA. This datasets collects lawyers' ratings of 43 different judges using 12 numeric variables: number of contacts of lawyer with judge (\texttt{CONT}), judicial integrity (\texttt{INTG}), demeanor (\texttt{DMNR}), diligence (\texttt{DILG}), case flow managing (\texttt{CFMG}), prompt decisions (\texttt{DECI}), preparation for trial (\texttt{PREP}), familiarity with law (\texttt{FAMI}), sound oral rulings (\texttt{ORAL}), sound written rulings (\texttt{WRIT}), physical ability (\texttt{PHYS}), and worthy of retention (\texttt{RTEN}). We are interested in ranking the judges by quality.

After standardizing the matrix $A \in \RR^{12 \times 43}$ (by ensuring each row had mean zero), we computed PCA of this dataset (Figure~\ref{subfig:PCA}), with $k = 2$. As evident in the plot, the data lacks directionality, which possibly means we have chosen $k$ to be too small. 

Next we used SMART to compute $20\%$-trimmed PCA on $A$ (Figure~\ref{subfig:T-PCA-Full}, discovered outliers plotted as red squares). After trimming 20\% of the dataset, it exhibited much greater directionality. In particular, the judges in the bottom right corner of Figure~\ref{subfig:T-PCA-Full} were rated poorly across all dimensions, while the judges in the upper left were rated highly across all dimensions. 
%The median ratings across the dataset were as follows: \texttt{CONT 7.3, INTG 8.1, DMNR 7.7, DILG 7.8, CFMG 7.6, DECI 7.7, PREP 7.7, FAMI 7.6, ORAL 7.5, WRIT 7.6, PHYS 8.1, RTEN 7.8}. There was no clear pattern among the outlying judges. Some were rated especially high, for example,  \texttt{CALLAHAN, R.J.} was rated \texttt{CONT 10.6, INTG 9, DMNR 8.9, DILG 8.7, CFMG 8.5, DECI 8.5, PREP 8.5, FAMI 8.5, ORAL 8.6, WRIT 8.4, PHYS 9.1, RTEN 9}, some were rated especially low, for example, \texttt{BRACKEN, J.J.} was rated \texttt{CONT 7.3, INTG 6.4, DMNR 4.3, DILG 6.5, CFMG 6, DECI 6.2, PREP 5.7, FAMI 5.7, ORAL 5.1, WRIT 5.3, PHYS 5.5, RTEN 4.8}, while others were rated close to the median in some respects and distant in others, for example, \texttt{DRISCOLL, P.J.} was rated \texttt{CONT 6.7, INTG 8.6, DMNR 8.2, DILG 6.8, CFMG 6.9, DECI 6.6, PREP 7.1, FAMI 7.3, ORAL 7.2, WRIT 7.2, PHYS 8.1, RTEN 7.7}.

We hypothesized that some of the 12 variables were uninformative for predicting the quality of a judge. For example, it is not clear how \texttt{CONT} relates to quality because it is not controlled by the judge, but may depend on the trial. Thus, we used SMART to compute 60\%-trimmed PCA on the transposed matrix $A^T \in \RR^{43 \times 12}$ and discovered the outlying categories \texttt{CONT}, \texttt{DMNR}, \texttt{INTG}, and \texttt{PHYS}. We removed these variables from the dataset, which resulted in a reduced data matrix $B \in \RR^{8 \times 43}$. Then we performed PCA on this new data matrix $B$ (Figure~\ref{subfig:PCA-removed}). Interestingly, some of the outliers found by 20\%-trimmed PCA on $A$, for example \texttt{BRACKEN, J.J} and \texttt{DRISCOLL, P.J.}, were removed from the center of the point cloud, making them easier to spot visually, while others no longer appeared to be outliers, for example, \texttt{CALLAHAN, R.J.}

The point cloud produced by standard PCA still lacked clear directionality. Thus, we used SMART to compute 20\%-trimmed PCA on $B$ (Figure~\ref{subfig:T-PCA-removed}, discovered outliers plotted as red squares). Figure~\ref{subfig:T-PCA-removed} shows that trimmed PCA now found a clear linear component of the data: the judges in the upper left hand are poorly rated, the judges in the middle of the figure are near the median, and the judges in the bottom right are highly rated. Compared to 20\%-trimmed PCA on $A$, some of the outliers persist, for example, \texttt{BRACKEN, J.J} and \texttt{DRISCOLL, P.J.}, while others cease to be outliers, for example, \texttt{CALLAHAN, R.J.} and \texttt{DANNEHY, J.F.} One hypothesis for why \texttt{DRISCOLL, P.J.} persists as an outlier is that he or she was rated low with respect to \texttt{DILG, CFMG, DECI} and \texttt{PREP}, but is still considered worthy of retention. One hypothesis for why \texttt{CALLAHAN, R.J.} was an outlier with respect to $A$ and not with respect to $B$ is that he or she received a high rating for \texttt{CONT}, $10.6$, while the mean and median for these ratings were $7.4$ and $7.3$.

\subsection{Robust homography estimation}

\begin{figure}[t!]
    \begin{subfigure}[t]{0.5\textwidth}
        \centering
        \includegraphics[scale=0.25]{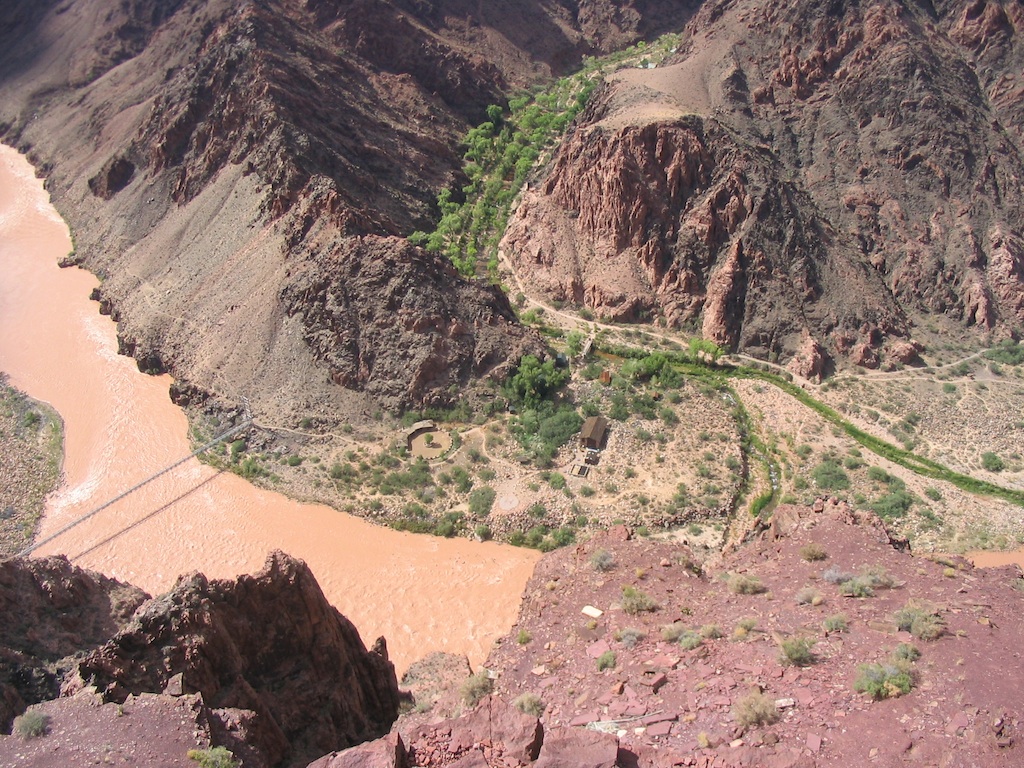} 
        \caption{Image 1}\label{subfig:river1}
    \end{subfigure}%
    ~ 
    \begin{subfigure}[t]{0.5\textwidth}
        \centering
        \includegraphics[scale=0.25]{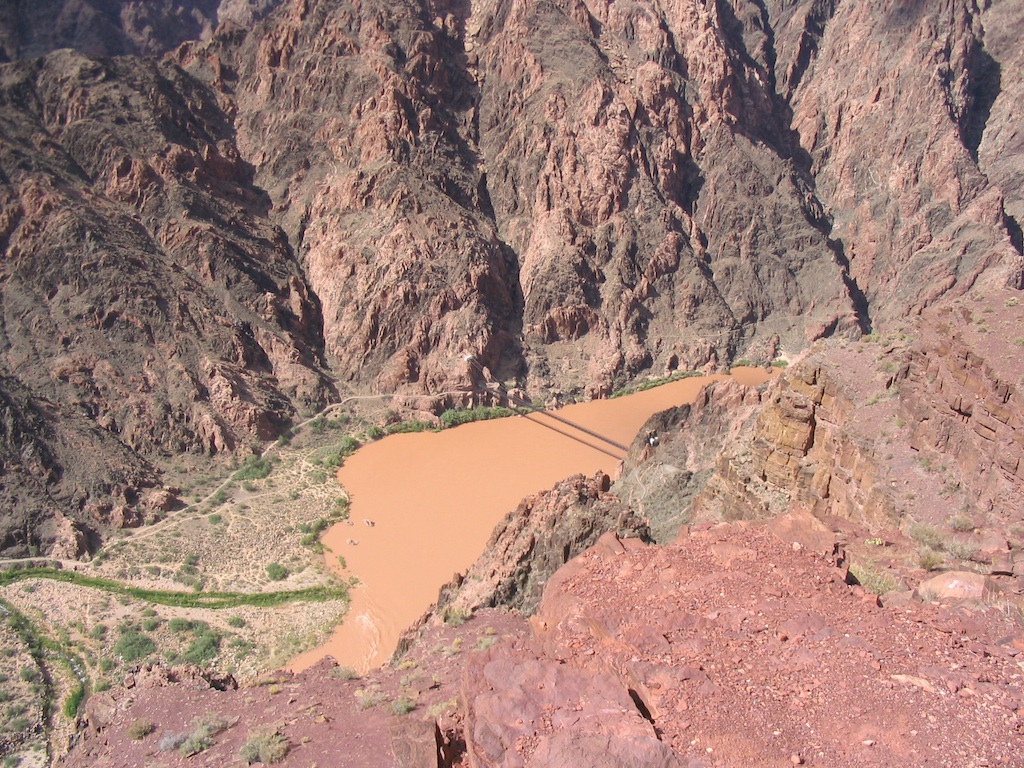} 
        \caption{Image 2}\label{subfig:river2}
    \end{subfigure}\\
        \begin{subfigure}[t]{1\textwidth}
        \centering
        \includegraphics[scale=.25]{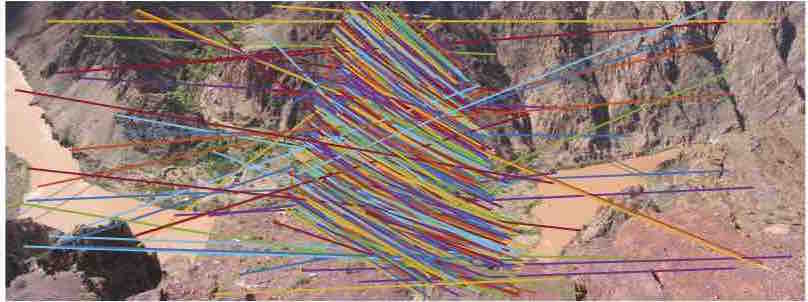} 
        \caption{Tentative matches from SIFT}\label{subfig:SIFT}
        \end{subfigure}\\
             \begin{subfigure}[t]{1\textwidth}
        \centering
        \includegraphics[scale=.25]{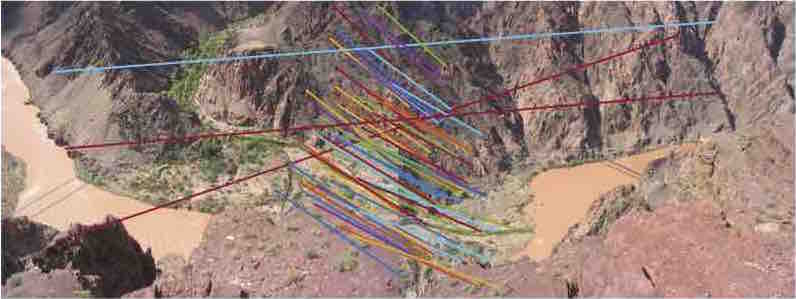} 
        \caption{Best 10\% matches found by SMART}\label{subfig:SIFT-SMART}
    \end{subfigure}
    %\caption{30\% contamination}
%\center
%\includegraphics[scale=0.6]{RateComp} 
\caption{\label{fig:river} 
%Panels~(a) and~(b) show two related images with some overlapping features. 
%Panel~(c) shows tentative point correspondences discovered by SIFT. Panel~(d) shows the trimmed correspondences obtained using SMART. 
All images and feature matches were generated with VLFeat~\citep{vedaldi08vlfeat}.
}
\end{figure}

\begin{figure}
        \centering
        \includegraphics[scale=0.2]{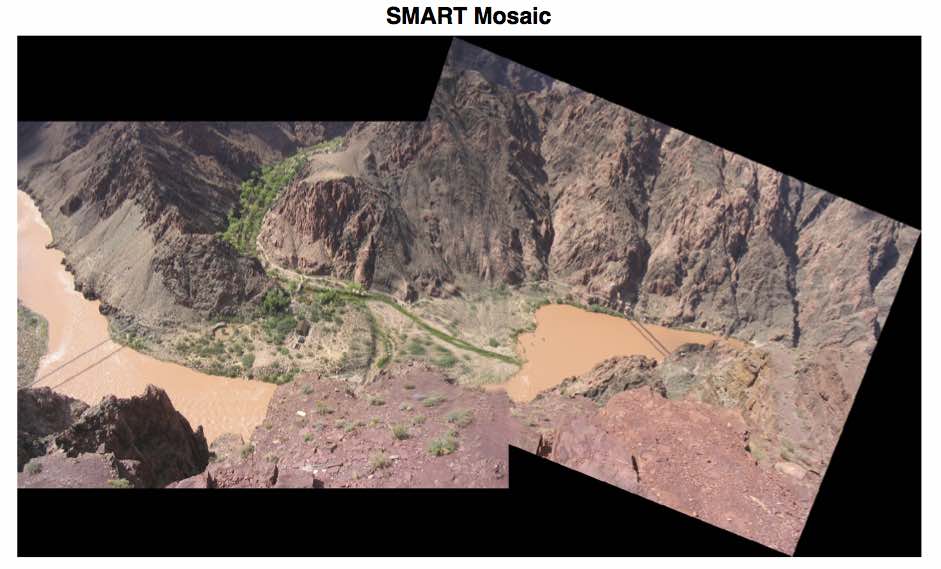} 
        \caption{\label{subfig:panorama} Final mosaic for images in Figure~\ref{fig:river} obtained using SMART.}
\end{figure}

Two images of the same scene, taken by a pin-hole camera, are related by a {\it homography} (see e.g.~\cite{hartley2003multiple,ma2012invitation}).
There exists a matrix $H \in \mathbb{R}^{3\times 3}$ so that given corresponding points $(u_1, v_1)$ in image $1$  and $(u_2, v_2)$ in image $2$, 
we have 
\begin{equation*}
H \begin{bmatrix} u_1 & v_1 &1\end{bmatrix}^T = \begin{bmatrix} u_2 & v_2 & 1\end{bmatrix}^T.
\end{equation*}
Given a set of point correspondences, we can determine $H$. Arranging corresponding sets of points into matrices $B_1$ and $B_2$, we can solve 
\begin{equation}
\label{eq:constr}
\min_{\|H\|_F = 1} \|H B_1 - B_2\|_F^2.  %\quad \Longrightarrow \quad \min_{\|h\|_2 = 1} \| (B_1^T\otimes I) h - \mathrm{vec}(B_2)\|^2.
\end{equation}
Given a perfect set of 4 point correspondences, the solution of~\eqref{eq:constr} is immediately obtained from the right singular vector, with singular value $0$, of a simple matrix $32$ by $8$ matrix~\cite{hartley2003multiple}. %foregoing any iterative algorithm to required to solve~\eqref{eq:constr}. 
This approach is known as direct linear transformation (DLT)~\citep{abdel1971direct}. 

The main challenge for homography estimation is finding a correct set of point correspondences. Potential point correspondences are generated with two steps. First, each image is scanned for visually distinctive points.  Those points deemed distinctive are assigned a vector (typically a 128 dimensional scale-invariant feature transform (SIFT)~\citep{lowe1999object} descriptor) that summarizes the neighborhood of the interest point. Second, by comparing descriptors between the images (typically with a nearest neighbors test) potential correspondences are generated between distinctive points.

After potential correspondences are generated, the random sample consensus (RANSAC) algorithm~\citep{fischler1981random} is used to remove erroneous correspondences. 
To do this, RANSAC repeatedly  selects a set of 4 points correspondences (uniformly at random), fits $h$ using the DLT procedure, and then estimates a 
consensus set, i.e. a set of point pairs $(m_1, m_2)$ whose errors $\|Hm_1 - m_2\|$ are smaller than a pre-defined threshold. 
Once the consensus set is large enough, the algorithm stops. 

\paragraph{Formulation.}  Given $n$ point correspondences, 
 rewriting~\eqref{eq:constr} as a sum over data points, and introducing weights, we solve  
\begin{equation}
\label{eq:SMARThomog}
\min_{w\in \Delta^h, \|H\|_F=1} \sum_{i=1}^n w_i \|Hb_{1,i} - b_{2,i}\|^2,
\end{equation}
which includes the nonconvex constraint $\|H\|_F = 1$.  We take the predicted number of inliers to be a small proportion of the data, say, 10\% or 20\%. 

%After solving~\eqref{eq:SMARThomog}, we do a refinement step to estimate the final homography. 
%We select the four best fitting correspondences (i.e., those with lowest objective values) and apply the DLT method as detailed above. 

\paragraph{Experiments.} 
We use~\eqref{eq:SMARThomog} to stitch together two overlapping images (shown in Figure~\ref{fig:river}). 
In our experiment, there are 627 point correspondences between the images (shown in Figure~\ref{subfig:SIFT}). 
Many of these correspondences are spurious. We trim away 90\% of the data using the SMART formulation~\eqref{eq:SMARThomog},
leaving only the correspondences shown in Figure~\ref{subfig:SIFT-SMART}. After solving~\eqref{eq:SMARThomog}, we do a refinement step to estimate the final homography. 
We select the four best fitting correspondences (i.e., those with lowest objective values) and apply the DLT method as detailed above. \\

Although SMART recovers a plausible mosaic, similar mosaics can also be recovered by RANSAC. However, for larger scale bundle adjustment problems, in which multiple images of the same scene are used to estimate several interconnected homographies, RANSAC becomes prohibitively slow. We expect SMART to perform well on these problems, but we leave them to future work.

\section{Conclusion}
We introduced the SMART algorithm for solving the nonconvex, nonsmooth problem~\eqref{eq:trimmedformulation}, which was motivated by the nonconvex trimming problem~\eqref{eq:trimmed}. SMART is the first stochastic gradient algorithm for fully nonconvex optimization that provably converges. Moreover, SMART scales better, by a factor of $n^{1/3}$, than all competing full gradient methods. In spite of the nonsmooth, nonconvex nature of~\eqref{eq:trimmedformulation}, we showed that SMART converges quickly, performs meaningful inference on contaminated datasets, and reliably detects outliers.

\small
\bibliographystyle{spmpscinat}      % mathematics and physical sciences
\bibliography{icml}   % name your BibTeX data base

\appendix

\section{Proof of Theorem~\ref{thm:smartconverges}}\label{app:proof}

\paragraph{Notation.} 
We will often repeat the following terms:
\begin{itemize}
\item \textbf{Conditional expectation $\EE_k$} For every $k \in \NN$, and every random variable $X$, we let $\EE_k \left[ X \right] = \EE\left[ X \mid \cF_k\right],$ where $\cF_k$ is defined as in Assumption~\ref{assumption:independence}.
\item \textbf{Stochastic Gradient Estimator.} For all $k \in \{0, \ldots, T-1\}$, we define an $\cH$-valued random variables $v^k$ with components
%\begin{align*}
$$
v^k := \frac{1}{b}\sum_{i \in I_k}(w_{i}^k \nabla f_{i}(x^k)- y_{i}^k) + \frac{1}{n} \sum_{i=1}^n y_{i}^k.
$$
%\end{align*}
\item \textbf{Full Update.} For all $k \in \{0,\ldots T-1\}$, we define a vector $\overline{z}^{k+1} \in \cH$ componentwise: 
\begin{align*}
\overline{z}^{k+1} &= \prox_{\gamma r_2}\left(x^k - \gamma v^k\right).
\end{align*}
%\item \textbf{Parametric Form of Dual Variables.} We explicitly define the $\cH$-valued random variables $\phi_{ij}^k \in \cH$ which generate the dual variables:
%\begin{align*}
%\left(\forall k \in \{0, \ldots, T\} \right), \left(\forall  i \in \{1, \ldots, n\}\right), \left(\forall j \in \{1, \ldots, m\}\right) \qquad\qquad y_{i,j}^k := \nabla_j f(\phi_{ij}^k).
%\end{align*}
%Note that $\phi_{ij}^k \in \{z^k \}_{j < k}$.
\item \textbf{The $\beta_{i}$ Factors.} Set 
%\begin{align*}
$$
\beta_{i} := \sqrt{1-\rho_{i}}\left(\frac{1}{\sqrt{q'}} -\sqrt{1-\rho_{i}}\right).
$$
%\end{align*}
\item \textbf{The $a$ Factor.} Set 
$$
a = \gamma L\sqrt{\frac{1}{n}\sum_{i=1}^n   \frac{q' (1+\epsilon_0)(B_i)^2}{2b \left(1 - \sqrt{q'(1-\rho_{i})}\right)^2}}.
$$ 
\item \textbf{The $\alpha_{i}$ Factors.} We let 
%\begin{align*}
$$
\alpha_{i} := \frac{q'\gamma(1+\epsilon_0)}{2ab}\sum_{t=0}^{\infty}\left[q'(1+\beta_{i})(1-\rho_{i})\right]^t = \frac{q'\gamma(1+\epsilon_0)}{2ab\left[ 1- q'(1+\beta_{i})(1-\rho_{i})\right]}.
$$
%\end{align*}
We use the property that $\alpha_i = \alpha_iq'(1+\beta_{i})(1-\rho_{i})+ \frac{q'\gamma(1+\epsilon_0)}{2ab}.$
\item \textbf{The Residuals.} For all $k \in \NN$, define
\begin{align*}
R_{\overline{w}}^k = \|\overline{w}^{k+1} - w^k\|^2; && R_{\overline{x}}^k &= \|\overline{x}^{k+1} - x^k\|^2; && R_{\overline{z}}^k = \|\overline{z}^{k+1} - x^k\|^2; \\
&& \Dk_i &= \|w_{i}^k\nabla f_i(x^k) - y_{i}^k\|^2. && 
\end{align*}
By our assumptions, $\Rwk$, $\Rxk$, and $\Dk_i$ are $\cF_k$-measurable. In contrast, $\Rzk$ is not necessarily $\cF_k$-measurable.
\end{itemize}

\paragraph{Parts~\ref{thm:smartconverges:item:decreaseinobjective} and~\ref{thm:smartconverges:item:convergencetostationarypoint}.}The supermartingale convergence theorem is our hammer for nailing down the effect of randomness in T-SMART:
\begin{theorem}[Supermartingale Convergence Theorem]\label{theorem:SCT}
Let $(\Omega, \cF, P)$ be a probability space. Let $\mathfrak{F} := \{\cF_k\}_{k \in \NN}$ be an increasing sequence of sub $\sigma$-algebras of $\cF$ such that $\cF_k \subseteq \cF_{k+1}$. Let $\{X_k\}_{k \in \NN}$ and $\{Y_k\}_{k \in \NN}$ be sequences of $[\xi, \infty)$-valued and $[0, \infty)$-valued random variables, respectively, such that for all $k \in \NN$, $X_k$ and $Y_k$ are $\cF_k$ measurable, and 
\begin{align}\label{eq:supermartingaleinequality}
(\forall k \in \NN) \qquad \EE\left[X_{k+1}\mid \cF_k\right] + Y_{k} \leq X_k.
\end{align}

Then $\sum_{k =0}^\infty Y_k < \infty \as$ and $X_k \as$ converges to a $[\xi, \infty)$-valued random variable.
\end{theorem}

In this proof, we show that~\eqref{eq:supermartingaleinequality} holds for the random variables (where $k \in \NN$)\footnote{The variable $X_k$ is clearly $\cF_k$-measurable. The variable $Y_k$ is $\cF_k$-measurable because of our assumptions on $\zeta_{1}$ and $\zeta_2$.}
\begin{align*}
 \;  X_{k} &= F(w^{k}, x^{k}) + \frac{1}{n}\sum_{i=1}^n \alpha_{i}  \Dk_i;\\
Y_k &= \frac{q'\gamma}{2\eta}\left\|\frac{\eta}{\gamma}\left(\overline{x}^{k+1} - x^k\right)\right\|^2+  \frac{q\tau}{2}\left\|\frac{1}{\tau}\left(\overline{w}^{k+1} - w^k\right)\right\|^2 + \frac{\epsilon_0 q'\gamma }{2abn}\sum_{i=1}^n \Dk_i. \numberthis\label{eq:supermartingale_definition}
\end{align*}
Then we apply Theorem~\ref{theorem:SCT} to show that $\sum_{k=0}^\infty Y_k < \infty \as$ and $X_{k} \as$ converges to a $[F(w^\ast, x^\ast), \infty)$-valued random variable $X_\ast$. 

Thus, there exists a full measure subset $\widetilde{\Omega} \subseteq \Omega$ such that the following hold: For all $\omega \in \widetilde{\Omega}$, the sequence $\{(w^{k}(\omega), x^{k}(\omega))\}_{k \in \NN}$ is bounded and 
\begin{enumerate}
\item Because $X_k \rightarrow X_\ast \as$ and $n^{-1}\sum_{i=1}^n\Dk_i \rightarrow 0$ as $k \rightarrow \infty$, we have $F(w^k(\omega), x^k(\omega)) \rightarrow X_\ast(\omega)$ as $k \rightarrow \infty$.
\item Because $\sum_{k=0}^\infty Y_k < \infty\as$, we have $\|\overline{x}^{k+1}(\omega) - x^k(\omega)\|^2 \rightarrow 0$ and $\|\overline{w}^{k+1}(\omega) - w^k(\omega)\|^2 \rightarrow 0$ as $k \rightarrow \infty$. 
\end{enumerate}

We use these limits to prove properties of convergent subsequences of T-SMART along the full measure set $\widetilde{\Omega}$.
\begin{lemma}\label{lem:stationary_point}
Let $\omega \in \widetilde{\Omega}$. Suppose that there exists an increasing sequence of indices $\{k_l\}_{l \in \NN} \subseteq \NN$ with the property that $(\overline{w}^{k_l+1}(\omega), \overline{x}^{k_l+1}(\omega)) \rightarrow (\overline{w}, \overline{x})$. Then $F(\overline{w}, \overline{x}) = X_\ast(\omega)$, the limit holds $F(\overline{w}^{k_l+1}(\omega), \overline{x}^{k_l+1}(\omega)) \rightarrow X_\ast(\omega) = F(\overline{w}, \overline{x})$, and there exists $g^{k_l} \in \partial_L F(\overline{w}^{k_l+1}, \overline{x}^{k_l+1})$ such that $g^{k_l} \rightarrow 0$ as $l \rightarrow \infty$. Therefore, $0 \in \partial_L F(\overline{w}, \overline{x})$.
\end{lemma}

Thus, Parts~\ref{thm:smartconverges:item:decreaseinobjective} and~\ref{thm:smartconverges:item:convergencetostationarypoint} follow as soon as we prove~\eqref{eq:supermartingaleinequality} for $X_k$ and $Y_k$. It turns out that Part~\ref{thm:smartconverges:item:sublinearrate} also follows from~\eqref{eq:supermartingaleinequality}.

\paragraph{Part~\ref{thm:smartconverges:item:sublinearrate}.}
If we apply the law of total expectation to~\eqref{eq:supermartingaleinequality}, we find that 
\begin{align*}
\min_{t = 0, \ldots, T}\EE\left[Y_t\right] \leq \frac{1}{T}\sum_{t=0}^T \EE\left[Y_t\right] \leq \frac{1}{T}\EE\left[ X_0 - X_{T+1}\right]\leq \frac{1}{T}\EE\left[F(w^0, x^0) - F(w^\ast, x^\ast)\right]
\end{align*} 
because $\sum_{i=1}^n \alpha_i\left\|w_i^0\nabla f_i(x^{0}) - y_{i}^{0}\right\|^2 = 0 $ and $F(w^T, x^T) \geq F(w^\ast, x^\ast)$. 

\paragraph{Lemmas Leading to~\eqref{eq:supermartingaleinequality}}
The proof of~\eqref{eq:supermartingaleinequality} requires four lemmas, whose proofs we defer for a moment. Though similar, the first lemma does not follow from~\cite[Lemma 2]{reddi2016fast}.
\begin{lemma}[Sufficient Decrease]\label{lem:sufficient_decrease}
For all $k\in \NN$, we have
\begin{align*}
&\EE_k\left[F(w^{k+1}, x^{k+1})\right] \leq F(w^k, z^k) + q'\EE_k\left[\dotp{\overline{z}^{k+1} - \overline{x}^{k+1},\frac{1}{n} \sum_{i=1}^n  w_i^k\nabla f_i(x^k)  - v^k}\right]  \\
&\hspace{0pt}+ q'\left[\frac{L}{n}\sum_{i=1}^n B_i  - \frac{\eta-1}{2\gamma}\right] \Rxk +q' \EE_k\left[\left[\frac{L}{n}\sum_{i=1}^n\frac{|w_i^k|}{2} - \frac{1}{2\gamma}\right] \Rzk\right] -\frac{\epsilon_0 q'\gamma }{2abn}\sum_{i=1}^n \Dk_i -  \frac{q}{2\tau}\Rwk.
\end{align*}
\end{lemma}

\begin{lemma}[Variance Bound]\label{lem:variancebound}
For all $k \in \NN$, we have
$$
\EE_k\left[ \left\|\frac{1}{n}\sum_{i=1}^n w_i^k\nabla f_i(x^k) - v^k\right\|^2\right]  \leq  \frac{1}{bn}\sum_{i=1}^n \Dk_i.%\\
$$
\end{lemma}

\begin{lemma}[Dual Variable Recursion]\label{lem:dual_variable}
For all $k \in \NN$, and $i \in \{1, \ldots, n\}$, we have
%\begin{align*}
$$
\EE_k\left[V^{k+1}_i \right] \leq q'\left(1+\frac{1 - \rho_i}{\beta_{i}}\right) (Lw_i^k)^2\EE_k\left[\Rzk\right] + q'(1-\rho_i)(1+\beta_{i}) \Dk_i.
$$
%\end{align*}
\end{lemma}

\begin{lemma}[$\alpha_{i}$ bound]\label{lem:c_bound}
The following bound holds: 
%\begin{align*}
$$
\frac{1}{n}\sum_{i=1}^n \left[\alpha_{i}(w_i^kL)^2\left( 1+ \frac{(1- \rho_{i} )}{\beta_{i}}\right)+ \frac{|w_i^k| L }{2}\right] \leq  \frac{1-2a}{2\gamma} .
$$
%\end{align*}
\end{lemma}

\paragraph{Proof of~\eqref{eq:supermartingaleinequality}} Using the variance bound, we bound the cross term from Lemma~\ref{lem:sufficient_decrease}:
\begin{align*}
 & \EE_k\left[\dotp{\overline{z}^{k+1} - \overline{x}^{k+1}, \frac{1}{n}\sum_{i=1}^nw_i^k\nabla f_i(x^k)   - v^k}\right]  \\
 &\leq \EE_k\left[ \frac{a}{2\gamma}\|\overline{z}^{k+1}- \overline{x}^{k+1}\|^2 + \frac{\gamma }{2a}\left\|\frac{1}{n}\sum_{i=1}^nw_i^k \nabla f_i(x^k)- v^k\right\|^2\right] 
%&\leq\EE_k\left[\frac{a}{2\gamma}\|\overline{z}^{k+1}- \overline{x}^{k+1}\|^2 \right]+ \frac{\gamma }{2abn}\sum_{i=1}^n \Dk_i \\
%&=  \sum_{j=1}^m\EE\left[\frac{1}{8\gamma_j}\|z_j^{k+1}- \overline{x}^{k+1}_j\|^2 \right]+ \EE\left[\sum_{j=1}^m 2\gamma_jq_j\| \nabla_j f(z^k) - u_j^k\|^2\right] \\
\leq \EE_k\left[\frac{a}{\gamma}\Rzk \right] + \frac{a}{\gamma}\Rxk + \frac{\gamma }{2abn}\sum_{i=1}^n \Dk_i, 
\end{align*}
where the first inequality follows from the Cauchy-Schwarz inequality and the bound $cd \leq c^2a/(2\gamma) +  d^2\gamma/2a$, and the last inequality follows from the bound $\|c +d\|^2 \leq 2\|c\|^2 + 2\|d\|^2$ and the $\cF_k$-measurability of $\Rxk = \|\overline{x}^{k+1}- x^{k}\|^2$.

Thus, the cross term bound taken together with Lemma~\ref{lem:sufficient_decrease} yields
\begin{align*}
\EE_k\left[F(w^{k+1}, x^{k+1})\right] &\leq F(w^k, x^k) +q' \left[\frac{L}{n}\sum_{i=1}^n B_i - \frac{(\eta - 1 - 2a)}{2\gamma} \right] \Rxk \\
&\hspace{5pt}+ q'\EE_k\left[\left[\frac{L}{n}\sum_{i=1}^n\frac{|w_i^k|}{2} - \frac{1 - 2a}{2\gamma}\right] \Rzk\right]  +\frac{q'\gamma }{2a bn}\sum_{i=1}^n \Dk_i - \frac{q}{2\tau}\Rwk.
\end{align*}
Therefore,
\begin{align*}
\EE_k\left[X_{k+1}\right] &\leq F(w^k, x^k) +\frac{1}{n} \sum_{i=1}^n \EE_k\left[\alpha_{i}  V^{k+1}_i\right] + q'\left[\frac{L}{n}\sum_{i=1}^n B_i  - \frac{(\eta - 1 - 2a)}{2\gamma} \right] \Rxk -\frac{q}{2\tau}\Rwk \\
&\hspace{5pt}+ q'\EE_k\left[\left[\frac{L}{n}\sum_{i=1}^n\frac{|w_i^k|}{2} - \frac{1 - 2a}{2\gamma}\right] \Rzk\right]+\frac{q' \gamma }{2abn}\sum_{i=1}^n \Dk_i.
\end{align*}
Using the dual variable recursion bound, we find that 
\begin{align*}
\sum_{i=1}^n \EE_k\left[\alpha_{i}  V_i^{k+1}\right] &\leq  \EE_k\left[\sum_{i=1}^n\alpha_{i}q'(w_i^kL)^2\left[ 1+ \frac{(1- \rho_{i})}{\beta_{i}}\right]\Rzk\right]  + \sum_{i=1}^n \alpha_{i}q'(1+\beta_i)(1-\rho_{i})\Dk_i.
\end{align*}
Thus, 
\begin{align*}
&\EE_k\left[X_{k+1}\right] \leq F(w^k,z^k) + \frac{1}{n}\sum_{i=1}^n \left[\alpha_{i}q'(1+\beta_{i})(1-\rho_{i} ) + \frac{\gamma q'(1+\epsilon_0)}{2ab} \right]\Dk_i\\
&\hspace{0pt} +  q'\EE_k\left[ \left[\frac{1}{n}\sum_{i=1}^n \left[\alpha_{i}(w_i^kL)^2\left( 1+ \frac{(1- \rho_{i} )}{\beta_{i}}\right)+ \frac{w_i^k L}{2}\right] - \frac{1-2a}{2\gamma} \right] \Rzk\right] \\
&\hspace{0pt}+ q'\left[\frac{L}{n}\sum_{i=1}^n B_i - \frac{(\eta - 1 - 2a)}{2\gamma} \right] \Rxk - \frac{q}{2\tau}\Rwk- \frac{\epsilon_0 q' \gamma}{2abn}\sum_{i=1}^n \Dk_i.\\
&\leq X_k -  \frac{q'\gamma}{2\eta} \left\|\frac{\eta}{\gamma}\left(\overline{x}^{k+1} - x^k\right)\right\|^2 -  \frac{q\tau}{2}\left\|\frac{1}{\tau}\left(\overline{w}^{k+1} - w^k\right)\right\|^2- \frac{\epsilon_0 q'\gamma }{2abn}\sum_{i=1}^n \Dk_i\\
&\leq X_k- Y_k, \numberthis\label{eq:thing_to_cite_in_linear}
\end{align*}
where the final inequalities follow from Lemma~\ref{lem:c_bound}, the definition of $\alpha_{i}$, and the identity
\begin{align*}
\eta = 2 \left(1+2a + 2\gamma\frac{L}{n}\sum_{i=1}^n B_i\right)
\implies \frac{\eta}{2\gamma} = \frac{(\eta - 1 - 2a)}{2\gamma} - \frac{L}{n}\sum_{i=1}^n B_i  .
\end{align*}  
\paragraph{Proofs of the Lemmas.}
\begin{proof}[of Lemma~\ref{lem:stationary_point}]
We first prove that $F(\overline{w}^{k_l+1}(\omega), \overline{x}^{k_l+1}(\omega)) \rightarrow F(\overline{w}, \overline{x})$, then we construct the subgradients. 

Because $\|\overline{x}^{k+1}(\omega) - x^k(\omega)\|^2 \rightarrow 0$ and $\|\overline{w}^{k+1}(\omega) - w^k(\omega)\|^2 \rightarrow 0$ as $k \rightarrow \infty$, it follows that $(w^{k_l}, x^{k_l}) \rightarrow (\overline{w}, \overline{x})$ as $l \rightarrow \infty$. Thus, by continuity, we have
$$
\lim_{l \rightarrow \infty} \frac{1}{n}\sum_{i=1}^n \overline{w}^{k_l+1}_i(\omega)f_i(\overline{x}^{k_l+1}(\omega)) = \frac{1}{n}\lim_{l \rightarrow \infty} \sum_{i=1}^n w_i^{k_l}(\omega)f_i(x^{k_l}(\omega)) = \frac{1}{n}\sum_{i=1}^n \overline{w}_i(\omega)f_i(\overline{x}(\omega)).
$$
Proving that $\lim_{l \rightarrow \infty} \left\{r_1(\overline{w}^{k_l+1}(\omega)) + r_2(\overline{x}^{k_l+1}(\omega))\right\} = \lim_{l \rightarrow \infty} \left\{r_1(w^{k_l}(\omega)) + r_2(x^{k_l}(\omega))\right\} = r_1(\overline{w}) + r_2(\overline{x})$ is a little subtler because $r_1$ and $r_2$ are not continuous, but merely lower-semicontinuous. 

Because $F(w^k(\omega), x^k(\omega)) \rightarrow X_\ast(\omega)$, we know the following limit exists:
$$
\lim_{l \rightarrow \infty} \left\{r_1(w^{k_l}(\omega)) + r_2(x^{k_l}(\omega))\right\} = X_\ast(\omega) - \frac{1}{n}\sum_{i=1}^n \overline{w}_i(\omega) f_i(\overline{x}(\omega)).
$$
Now we focus on proving that $r_1(\overline{w}^{k_l+1}(\omega)) + r_2(\overline{x}^{k_l+1}(\omega))$ has the same limit as $l \rightarrow \infty$.

First, 
\begin{align*}
r_1(\overline{w}^{k_l+1}(\omega)) &\leq r_1(w^{k_l}(\omega)) + \frac{1}{n}\sum_{i=1}^n (w_i^{k_l}(\omega)  - \overline{w}_i^{k_l+1}(\omega) ) f_i(x^{k_l}(\omega)) - \frac{1}{2\tau}\|\overline{w}^{k_l + 1}(\omega) - w^{k_l}(\omega)\|^2 \\
r_2(\overline{x}^{k_l+1}(\omega)) &\leq r_2(x^{k_l}(\omega)) + \left\langle\frac{1}{n}\sum_{i=1}^n w_i^{k_l}(\omega)\nabla f_i(x^{k_l}(\omega)) , x^{k_l}(\omega) - \overline{x}^{k_l+1}(\omega)\right\rangle
%\\ &\hspace{20pt}
- \frac{\eta}{2\gamma}\|\overline{x}^{k_l+1}(\omega) - x^{k_l}(\omega)\|^2.
\end{align*}
Taking $\liminf$ of both sides as $l \rightarrow \infty$, we find that $$\limsup_{l \rightarrow \infty} \left\{r_1(\overline{w}^{k_l+1}(\omega)) + r_2(\overline{x}^{k_l+1}(\omega))\right\} \leq \lim_{l \rightarrow \infty} \left\{r_1(w^{k_l}(\omega)) + r_2(x^{k_l}(\omega))\right\},$$
where we have implicitly used that $\left\{(x^{k_l}(\omega), w^{k_l}(\omega))\right\}_{l \in \NN}$ is bounded. 

Second, for all $k \in \NN$, define
\begin{align*}
d(k, w) = \max\left\{\{t < k \mid j_t(\omega) = 1\} \cup\{-1\}\right\} && \text{and} && d(k, x) = \max\left\{\{t < k \mid j_t(\omega) = 2\} \cup\{-1\}\right\}.
\end{align*}
Without loss of generality, we now assume that $k_0$ is large enough that $d(k_0, w) > 0$ and $d(k_0, x) > 0$.
\begin{align*}
  r_1(w^{k_l}(\omega)) &\leq r_1(\overline{w}^{k_l+1}(\omega)) + \frac{1}{n}\sum_{i=1}^n (\overline{w}_i^{k_l+1}(\omega) - w_i^{k_l}(\omega) ) f_i(x^{d(k_l, w)}(\omega))\\
 &\hspace{5pt} + \frac{1}{2\tau}\|\overline{w}^{k_l + 1}(\omega) - w^{d(k_l, w)}(\omega)\|^2 -  \frac{1}{2\tau}\|w^{k_l }(\omega) - w^{d(k_l, w)}(\omega)\|^2. \\
 &= r_1(\overline{w}^{k_l+1}(\omega)) + \frac{1}{n}\sum_{i=1}^n (\overline{w}_i^{k_l+1}(\omega) - w_i^{k_l}(\omega)) f_i(x^{d(k_l, w)}(\omega)) \\
 &\hspace{5pt} + \frac{1}{2\tau}\left[2\dotp{\overline{w}^{k_l + 1}(\omega) - w^{d(k_l, w)}(\omega), \overline{w}^{k_l + 1}(\omega) - w^{k_l}(\omega)} - \|\overline{w}^{k_l + 1}(\omega) - w^{k_l }(\omega)\|^2\right],
\end{align*}
and similarly for $r_2$, we have
\begin{align*}
r_2(x^{k_l}(\omega)) &\leq r_2(\overline{x}^{k_l+1}(\omega))+ \dotp{v^{d(k_l, x)}(\omega), \overline{x}^{k_l+1}(\omega) - x^{k_l}(\omega) }\\
&\hspace{20pt} + \frac{\eta}{2\gamma}\left[\dotp{\overline{x}^{k_l + 1}(\omega) - x^{d(k_l, x)}(\omega), \overline{x}^{k_l + 1}(\omega) - x^{k_l}(\omega)} - \|\overline{x}^{k_l+1}(\omega) - x^{k_l}(\omega)\|^2\right].
\end{align*}
Thus, because $w^{d(k_l, w)}(\omega), x^{d(k_l, x)}(\omega), v^{d(k_l, x)}(\omega)$ are all bounded, by taking $\liminf$ of both sides as $l \rightarrow \infty$, we find that 
$$\lim_{l \rightarrow \infty} \left\{r_1(w^{k_l}(\omega)) + r_2(x^{k_l}(\omega))\right\} \leq \liminf_{l \rightarrow \infty} \left\{r_1(\overline{w}^{k_l+1}(\omega)) + r_2(\overline{x}^{k_l+1}(\omega))\right\}.$$

Therefore, we've shown that $r_1(\overline{w}^{k_l+1}(\omega)) + r_2(\overline{x}^{k_l+1}(\omega))$ and $r_1(w^{k_l}(\omega)) + r_2(x^{k_l}(\omega))$ have the same limit at $l \rightarrow \infty.$ Now we show that $\lim_{l \rightarrow \infty} \left\{r_1(w^{k_l}(\omega)) + r_2(x^{k_l}(\omega))\right\} = r_1(\overline{w}) + r_2(\overline{x})$.

By the lower-semicontinuity of $r_1$ and $r_2$, we have
$
\lim_{l \rightarrow \infty} \left\{r_1(w^{k_l}(\omega)) +  r_2(x^{k_l}(\omega)) \right\} \geq r_1(\overline{w}) +  r_2(\overline{x}).
$
In addition, because $w^{k_l}(\omega)$ and $x^{k_l}(\omega)$ are proximal points, we have  
\begin{align*}
r_1(w^{k_l}(\omega)) &\leq r_1(\overline{w}) + \frac{1}{n}\sum_{i=1}^n (\overline{w}_i(\omega) - w_i^{k_l}(\omega))f_i(x^{d(k_l, j)}(\omega)) \\
 &\hspace{5pt} + \frac{1}{2\tau}\left[\dotp{\overline{w}- w^{d(k_l, w)}(\omega), \overline{w}- w^{k_l}(\omega)} - \|\overline{w} -  w^{k_l}(\omega)\|^2\right] \\
 r_2(x^{k_l}(\omega)) &\leq r_2(\overline{x})+ \dotp{v^{d(k_l, x)}(\omega), \overline{x} - x^{k_l}(\omega) }\\
&\hspace{20pt} + \frac{\eta}{2\gamma}\left[\dotp{\overline{x} - x^{d(k_l, x)}(\omega), \overline{x}- x^{k_l}(\omega)} - \|\overline{x} - x^{k_l}(\omega)\|^2\right].
\end{align*}
Therefore, by arguments similar to those already employed above, we find that 
$$
\lim_{l \rightarrow \infty} \left\{r_1(w^{k_l}(\omega)) +  r_2(x^{k_l}(\omega)) \right\} \leq r_1(\overline{w}) +  r_2(\overline{x}).
$$
Thus, $\lim_{l \rightarrow \infty} \left\{r_1(w^{k_l}(\omega)) + r_2(x^{k_l}(\omega))\right\} = r_1(\overline{w}) + r_2(\overline{x})$.

Therefore, by taking all these limits together we have shown that $\lim_{l \rightarrow \infty} F(\overline{w}^{k_l+1}, \overline{x}^{k_l+1}) = \lim_{l \rightarrow \infty} F(w^{k_l}, x^{k_l}) = F(\overline{w}, \overline{x})$. Now we construct the subgradient $g^{k_l} \in \partial_L F(\overline{w}^{k_l+1}, \overline{x}^{k_l+1})$.

By definition of the proximal operator, we have
\begin{align*}
\frac{1}{\tau}\left(w^{k_l} - \overline{w}^{k_l+1}\right) &\in \frac{1}{n}(f_1(x^{k_l}), \ldots, f_n(x^{k_l})) + \partial_L r_1(\overline{w}^{k_l+1}); \\
\frac{\eta}{\gamma}\left(x^{k_l} - \overline{x}^{k_l+1}\right) &\in \frac{1}{n}\sum_{i=1}^n w_i^{k_l} \nabla f_i(x^{k_l})+ \partial_L r_2(\overline{x}^{k_l+1}).
\end{align*}
Then we let
$$
g^{k_l} = \begin{bmatrix}
\frac{1}{\tau}\left(w^{k_l} - \overline{w}^{k_l+1}\right) + \frac{1}{n}(f_1(\overline{x}^{k_l+1}) - f_1(x^{k_l}), \ldots, f_n(\overline{x}^{k_l+1})  - f_n(x^{k_l}))\\
\frac{\eta}{\gamma}\left(x^{k_l} - \overline{x}^{k_l+1}\right) +   \frac{1}{n}\sum_{i=1}^n\overline{w}_i^{k_l+1} \nabla f_i(\overline{x}^{k_l + 1}) - \frac{1}{n}\sum_{i=1}^n w_i^{k_l}\nabla f_i(x^{k_l})
\end{bmatrix} \in \partial_L F(\overline{w}^{k_l+1}, \overline{x}^{k_l+1}).
$$
By the limits $\|\overline{x}^{k+1}(\omega) - x^k(\omega)\|^2 \rightarrow 0$ and $\|\overline{w}^{k+1}(\omega) - w^k(\omega)\|^2 \rightarrow 0$ as $k \rightarrow \infty$ and by continuity, we find that $g^{k_l} \rightarrow 0$ as $l \rightarrow \infty$. 
By the definition of the limiting subdifferential~\cite[Definition 8.3]{RTRW}, it follows that $0 \in \partial_L F(\overline{w}, \overline{x})$.
\end{proof}

\begin{proof}[of Lemma~\ref{lem:sufficient_decrease}]
We use the standard descent Lemma, found in~\cite[Lemma 1.2.3]{opac-b1104789}, several times throughout the proof. 

The result follows by constructing three bounds and adding them together. The first bound: for all $i \in \{1, \ldots, n\}$, we have
\begin{align*}
w_i^k f_i(\overline{x}^{k+1}) &\leq w_i^k f_i(x^k) + \dotp{\overline{x}^{k+1} - x^k,w_i^k\nabla f_i(x^k) } + \frac{|w_i^k|L}{2}\Rxk;\\
r_2(\overline{x}^{k+1}) &\leq r_2(x^k) +\dotp{x^k - \overline{x}^{k+1}, \frac{1}{n}\sum_{i=1}^n w_i^{k}\nabla f_i(x^k)} - \frac{\eta}{2\gamma}\Rxk,
\end{align*}
which implies that
\begin{align*}
 &\frac{1}{n} \sum_{i=1}^n w_i^{k}f_i(\overline{x}^{k+1}) + r_1(\overline{x}^{k+1}) \leq \frac{1}{n} \sum_{i=1}^n w_i^{k} f_i(x^k) + r_1(x^{k}) + \left[\frac{L}{n}\sum_{i=1}^n \frac{|w_i^k|}{2} - \frac{\eta}{2\gamma}\right]\Rxk. \numberthis\label{eq:bound1}
\end{align*}

The second bound: For all $i \in \{1, \ldots, n\}$, we have 
\begin{align*}
w_i^k  f_i(\overline{z}^{k+1}) &\leq w_i^k f_i(x^{k}) + \dotp{\overline{z}^{k+1} - x^{k},w_i^k\nabla f_i(x^{k})} + \frac{|w_i^k|L}{2}\Rzk;  \text{ and}\\
w_i^k f_i(x^k) &\leq  w_i^k f_i(\overline{x}^{k+1}) + \dotp{x^k - \overline{x}^{k+1} , w_i^k\nabla f_i(x^{k})} + \frac{|w_i^k|L}{2}\Rxk,
\end{align*}
where the second bound follows from the inequality~\cite[Lemma 1.2.3]{opac-b1104789}. Adding these bounds together, we obtain
\begin{align*}
w_i^k f_i(\overline{z}^{k+1}) \leq w_i^k f_i(\overline{x}^{k+1}) + \dotp{\overline{z}^{k+1} - \overline{x}^{k+1}, w_i^k\nabla f_i(x^k)} + \frac{|w_i^k|L}{2}\Rzk+ \frac{|w_i^k|L}{2}\Rxk.
\end{align*}
Then, by the definition of the proximal operator, 
%\begin{align*}
%r_j(w_j^k) \leq r_j(z_j^k) + \dotp{z_j^k - w_j^k, v_j^k} - \frac{q_jm}{2\gamma}\|w_j^k - z_j^k\|^2.
%\end{align*}
\begin{align*}
r_2(\overline{z}^{k+1}) \leq r_2(\overline{x}^{k+1}) + \dotp{\overline{x}^{k+1} - \overline{z}^{k+1}, v^{k}} + \frac{1}{2\gamma}\Rxk - \frac{1}{2\gamma}\Rzk.
\end{align*}
Thus, 
\begin{align*}
 &\frac{1}{n} \sum_{i=1}^n w_i^{k}  f_i(\overline{z}^{k+1}) + r_1(\overline{z}^{k+1}) \leq  \frac{1}{n} \sum_{i=1}^n w_i^{k} f_i(\overline{x}^{k+1}) + r_1(\overline{x}^{k+1}) + \left[\frac{L}{n}\sum_{i=1}^n \frac{|w_i^k|}{2} - \frac{1}{2\gamma} \right] \Rzk \\
 &\hspace{5pt}+\left[\frac{L}{n}\sum_{i=1}^n \frac{|w_i^k|}{2}+ \frac{1}{2\gamma}\right] \Rxk + \dotp{ \overline{z}^{k+1} - \overline{x}^{k+1}, \frac{1}{n}\sum_{i=1}^n w_i^k \nabla f_i(x^k)  - v^{k} }.\numberthis\label{eq:bound2}
\end{align*}
Thus, by adding~\eqref{eq:bound1} and~\eqref{eq:bound2}, we have
\begin{align*}
& \frac{1}{n} \sum_{i=1}^n w_i^{k}  f_i(\overline{z}^{k+1}) + r_1(\overline{z}^{k+1}) \leq  \frac{1}{n} \sum_{i=1}^n w_i^{k}f_i(x^{k}) + r_1(x^{k}) + \left[\frac{L}{n}\sum_{i=1}^n \frac{|w_i^k|}{2} - \frac{1}{2\gamma} \right] \Rzk \\
 &\hspace{5pt}+\left[\frac{L}{n}\sum_{i=1}^n |w_i^k|- \frac{\eta - 1}{2\gamma}\right] \Rxk + \dotp{ \overline{z}^{k+1} - \overline{x}^{k+1}, \frac{1}{n}\sum_{i=1}^nw_i^k \nabla f_i(x^k)  - v^{k} }.\numberthis\label{eq:bound3}
\end{align*}

The third bound: we have
\begin{align*}
\frac{1}{n}\sum_{i=1}^n \overline{w}^{k+1}_i f_i(x^k)& = \frac{1}{n}\sum_{i=1}^n w_i^k  f_i(x^k) + \dotp{\overline{w}^{k+1} - w^k, \frac{1}{n}(f_1(x^k), \ldots, f_n(x^k))};\\
r_1(\overline{w}^{k+1}) &\leq r_1(w^k) + \dotp{w^k - \overline{w}^{k+1},  \frac{1}{n}(f_1(x^k), \ldots, f_n(x^k))} - \frac{1}{2\tau} \Rwk,
\end{align*}
which implies that
\begin{align*}
\frac{1}{n}\sum_{i=1}^n \overline{w}^{k+1}_i f_i(x^k) + r_1(\overline{w}^{k+1}) \leq \frac{1}{n}\sum_{i=1}^n w^{k}_i f_i(x^k) + r_1(w^{k}) - \frac{1}{2\tau} \Rwk.
\end{align*}
Therefore, we find that
\begin{align*}
&\EE_k\left[\frac{1}{n}\sum_{i=1}^n w_i^{k+1}f_i(x^{k+1}) + r_1(w^{k+1}) + r_2(x^{k+1})\right] \\
&\leq \frac{q}{n}\sum_{i=1}^n \overline{w}_i^{k+1} f_i(x^k) + r_1(\overline{w}^{k+1}) + r_2(x^{k}) + (1-q)\EE_k\left[\frac{1}{n}\sum_{i=1}^n w_i^{k}f_i(\overline{z}^{k+1}) + r_1(w^k) + r_2(\overline{z}^{k+1} ) \right] \\
&\leq \frac{1}{n}\sum_{i=1}^n  w_i^{k} f_i(x^{k}) + r_1(w^{k}) + r_2(x^{k}) + q'\left[\frac{L}{n}\sum_{i=1}^n\frac{B_i}{2} - \frac{\eta-1}{2\gamma}\right] \Rxk  -  \frac{ q}{2\tau}\Rwk\\
&\hspace{20pt} +q'\EE_k\left[ \left[\frac{L}{n}\sum_{i=1}^n\frac{|w_i^k|}{2} - \frac{1}{2\gamma}\right] \Rzk\right] + q'\EE_k\left[\dotp{\overline{z}^{k+1} - \overline{x}^{k+1}, \frac{1}{n}\sum_{i=1}^n w_i^k\nabla f_i(x^k)   - v^k}\right].\end{align*}
\end{proof}

\begin{proof}[of Lemma~\ref{lem:variancebound}]
 For all $i \in \{1, \ldots, n\}$, define $\xi_i^k := w_i^k\nabla f_i(x^k) - y_i^k$. Then we have
\begin{align*}
 \EE_{k, l \sim \text{Unif}[n]} \left[\xi_l^k \right] :=  \EE_{l \sim \text{Unif}[n]} \left[\xi_l^k  \mid \cF_k\right]  &=  \frac{1}{n}\sum_{i=1}^nw_i^k\nabla f_i(x^k) - \frac{1}{n} \sum_{i=1}^n y_{i}^k,
\end{align*}
we find that
\begin{align*}
&\EE_k\left[ \left\|\frac{1}{n}\sum_{i=1}^n w_i^k\nabla f_i(x^k) - v^k\right\|^2\right] \\
&= \EE_k \left[\left\|\frac{1}{b}\sum_{j \in I_s} \left[ (w_j^k \nabla f_j(x^k) - y_j^k) - \left( \frac{1}{n}\sum_{i=1}^nw_i^k\nabla f_i(x^k) - \frac{1}{n} \sum_{i=1}^n y_{i}^k\right)\right]\right\|^2\right] \\
&= \EE_k \left[\left\|\frac{1}{b}\sum_{j \in I_s} \left[\xi_j^k - \EE_{k, l \sim \text{Unif}[n]}\left[\xi_l^k \right]\right]\right\|^2 \right]\\
&= \EE_k \left[\frac{1}{b^2}\sum_{j \in I_s} \left\|\xi_j^k - \EE_{k, l \sim \text{Unif}[n]}\left[\xi_l^k \right]\right\|^2\right] \\
&= \EE_k \left[\frac{1}{b^2}\sum_{j \in I_s} \left(\left\|\xi_j^k\right\|^2 - \left\|\EE_{k, l \sim \text{Unif}[n]}\left[\xi_l^k \right]\right\|^2\right)\right] \\
&= \frac{1}{bn}\sum_{i =1}^n \left\|\xi_i^k\right\|^2 - \frac{1}{b}\left\| \frac{1}{n}\sum_{i=1}^nw_i^k\nabla f_i(x^k) - \frac{1}{n} \sum_{i=1}^n y_{i}^k\right\|^2 \\
&\leq \frac{1}{bn}\sum_{i=1}^n \Dk_i.
\end{align*}
where the third equality follows because $\sum_{j \in I_s} \left[\xi_j^k - \EE_{k, l \sim \text{Unif}[n]}\left[\xi_l^k \right]\right]$ is the sum of independent, zero mean random vectors. 
\end{proof}

\begin{proof}[of Lemma~\ref{lem:dual_variable}]
Set $q' = (1-q)$. Then
\begin{align*}
&\EE_k\left[V^{k+1}_i\right] = \EE_k\left[ \left\|w_{i}^{k+1}\nabla f_i(x^{k+1}) - y_{i}^{k+1}\right\|^2\right] \\
&= q \left\|\overline{w}_i^{k+1}\nabla f_i(x^{k}) - \overline{w}_i^{k+1}\nabla f_i(x^k)\right\|^2+ q'\rho_i\EE_k\left[ \left\|w_i^{k}\nabla f_i(\overline{z}^{k+1})-w_i^{k} \nabla f_i(x^{k})\right\|^2\right] \\
&\hspace{20pt}+  q'(1-\rho_i)\EE_k\left[ \left\|w_i^{k}\nabla f_i(\overline{z}^{k+1})- y_i^k\right\|^2\right]  \\
&\leq q'(\rho_i + (1-\rho_i)(1+\beta_i^{-1}))\EE_k\left[ \left\|w_i^{k}\nabla f_i(\overline{z}^{k+1}) - w_i^{k}\nabla f_i(x^{k})\right\|^2\right]  + q'(1-\rho_{i})(1+\beta_{i}) \Dk_i \\
&\leq q'\left(1+\frac{1 - \rho_i}{\beta_{i}}\right)\EE_k\left[ (L_{i}w_i^k)^2\Rzk\right] + q'(1-\rho_i)(1+\beta_{i}) \Dk_i,
\end{align*}
where the first inequality follows from the inequality $\|a + b\|^2 \leq (1+\beta_{i}^{-1})\|a\|^2 + (1+\beta_{i})\|b\|^2$.
\end{proof}

\begin{proof}[of Lemma~\ref{lem:c_bound}]
Let $i \in \{1, \ldots, n\}$ such that $\rho_i \neq 1$. Recall that 
$
\beta_{i} := \sqrt{1-\rho_{i}}\left((q')^{-1/2} -\sqrt{1-\rho_{i}}\right).
$
Define $\theta_{i} := 1 - q'(1 + \beta_i- \rho_i) = 1 -  \sqrt{q'(1-\rho_i)}$ and note that 
$
q'(1+\beta_{i})(1-\rho_{i} ) \leq q'(1 + \beta_{i} - \rho_{i})  =  1- \theta_{i}.
$
Thus, we have
\begin{align*}
\alpha_{i} \leq \frac{q'\gamma(1+\epsilon_0)}{2ab} \sum_{t= 0}^{\infty} \left[(1-\theta_{i})\right]^t &=  \frac{q'\gamma(1+\epsilon_0) }{2ab}\frac{1}{\theta_i}.% \leq  \frac{q'\gamma(1+\epsilon_0) }{2ab\left(1 - \sqrt{q'(1-\rho_{i})}\right)}.
\end{align*}

With this bound in hand, we find that  
$$
\alpha_{i} \left[ 1+ \frac{(1- \rho_{i})}{\beta_{i}}\right]  \leq  \frac{q'\gamma(1+\epsilon_0) }{2ab\theta_i^2}.%\left[  \frac{1}{\theta_i}\right] = \frac{q'\gamma(1+\epsilon_0) }{2a b\left(1 - \sqrt{q'(1-\rho_{i}})\right)^2}.
$$
Therefore, because
$a = \gamma\sqrt{\frac{1}{n}\sum_{i=1}^n   \frac{q' (1+\epsilon_0)(B_iL_{i})^2}{2 b\theta_i^2}},  
$
we have 
\begin{align*}
 \frac{1}{n}\sum_{i=1}^n \left[q'\alpha_{i}(Lw_i^k)^2\left( 1+ \frac{(1- \rho_{i} )}{\beta_{i}}\right)+ \frac{|w_i^k| L}{2}\right]  + \frac{a}{\gamma} &\leq\frac{1}{n}\left[\frac{\gamma}{a}\sum_{i=1}^n   \frac{q'(1+\epsilon_0) (B_i L)^2}{2 b\theta_i^2}
+ \frac{B_i L}{2}\right]  + \frac{a}{\gamma} \\
&= 2L\sqrt{\frac{1}{n}\sum_{i=1}^n   \frac{q'(1+\epsilon_0) B_i^2}{2 b\theta_i^2}} + \frac{L}{n}\sum_{i=1}^n \frac{B_i }{2} \leq \frac{1}{2\gamma},
\end{align*}
where the last inequality holds by assumption.
\end{proof}

\subsection{Proof of Corollary~\ref{cor:sublinear}}\label{app:sublinear}

Our choice of $\tau$ guarantees that $\frac{q\tau}{2} = \frac{q'\gamma}{2\eta}$. Thus, from Part~\ref{thm:smartconverges:item:sublinearrate} of Theorem~\ref{thm:smartconverges}, it is clear that SMART achieves accuracy $\varepsilon$ after $O\left(\frac{2\eta}{q'\gamma\varepsilon}\right)$ iterations. We estimate this ratio below.

Because $D_k \equiv I_k$, we find that $\rho_i = P(i  \in D_k) = 1 - (1-1/n)^b$. Thus, with  
$$
a' = \sqrt{\frac{1}{n}\sum_{i=1}^n   \frac{q' (1+\epsilon_0)(B_iL_{i})^2}{2 b\left(1 - \sqrt{q'(1-\rho_{i})}\right)^2}},  
$$
and $\zeta = 1-1/n$, we have $\gamma = (4a' + \frac{L}{n}\sum_{i=1}^nB_i)^{-1}$ and $\eta = 2+ 4\gamma \left(a' + \frac{L}{n}\sum_{i=1}^nB_i\right),$ and 
\begin{align*}
&\frac{\gamma}{2\eta} = \frac{1}{2\eta/\gamma} = \frac{1}{4(1/\gamma + 2a' + \frac{2L}{n}\sum_{i=1}^nB_i)} \\
&=  \frac{1}{4(4 a' + \frac{L}{n}\sum_{i=1}^nB_i  + 2a' + \frac{2L}{n}\sum_{i=1}^nB_i)} = \frac{1}{4L\left( \frac{6\sqrt{\frac{1}{n}\sum_{i=1}^n q' (1+\epsilon_0)B_i^2}}{\sqrt{2b}\left(1 - \sqrt{q'\zeta^b}\right)} + \frac{3}{n}\sum_{i=1}^n B_i\right) } \\
&= \frac{\sqrt{2b}\left(1 - \sqrt{q'\zeta^b}\right)}{4L\left( 6\sqrt{\frac{1}{n}\sum_{i=1}^n q' (1+\epsilon_0)B_i^2} + \sqrt{2b}\left(1 - \sqrt{q'\zeta^b}\right)\frac{3}{n} \sum_{i=1}^nB_i\right) } = \Omega\left(\frac{b^{3/2}}{n}\right),
\end{align*}
where we use the bound: $1 - \sqrt{q'\zeta^b} = 1 - \zeta^{(b+1)/2} \geq \frac{b+1}{4n}.$ Therefore, SMART achieves accuracy $\varepsilon$ in at most $O\left(\frac{2\eta}{q'\gamma\varepsilon} \right) = O\left(\frac{n}{b^{3/2}\varepsilon}\right)$
iterations. 

To initialize properly, SMART requires $n$ gradient evaluations. Then, on average, the $w$ variables will be updated once every $n$ steps, and each of those updates requires $n$ function evaluations, $n$ gradient evaluations, and 1 evaluation of $\prox_{\tau r_1}$. Thus, to reach accuracy $\varepsilon$, SMART requires on average at most $O(n/(b^{3/2}\varepsilon)) (1/n) n = O(n/(b^{3/2}\varepsilon))$ function evaluations and $O\left(n/(b^{3/2}\varepsilon)\right) (1/n) = O(1/(b^{3/2}\varepsilon))$ evaluations of $\prox_{\tau r_1}$. 

Similarly, the $x$ variables are updated every $1/(1-1/n) = O(1)$ iterations, and each update requires takes $b$ gradient evaluations, and $1$ evaluation of $\prox_{\gamma r_1}$. Thus, to reach accuracy $\varepsilon$, SMART requires at most 
$$
n + O(n/(b^{3/2}\varepsilon)) O(1) + O(n/(b^{3/2}\varepsilon)) O(b) = O(n + n/(b^{3/2}\varepsilon) + n/(b^{1/2}\varepsilon))
$$ 
gradient evaluations and $O(n/(b^{3/2}\varepsilon))$ evaluations of $\prox_{\gamma r_2}$.

\subsection{Proof of Corollary~\ref{cor:sublinear2}}\label{app:sublinear2}

The proof of this Corollary follows the exact same logic as the proof of Corollary~\ref{cor:sublinear}, up to the equation (with $\zeta = 1-1/n$)
\begin{align*}
&\frac{\gamma}{2\eta} = \frac{1}{2L}\left( \frac{6\sqrt{\frac{1}{n}\sum_{i=1}^n \zeta^b(1+\epsilon_0)B_i^2}}{\sqrt{2b}\left(1 - \sqrt{\zeta^b}\right)} + \frac{3}{n}\sum_{i=1}^n B_i\right)^{-1}\\
&= \frac{\sqrt{2b}\left(1 - \sqrt{\zeta^b}\right)}{2L\left( 6\sqrt{\frac{1}{n}\sum_{i=1}^n \zeta^b(1+\epsilon_0)B_i^2} + \sqrt{2b}\left(1 - \sqrt{\zeta^b}\right)\frac{3}{n} \sum_{i=1}^nB_i\right)} = \Omega\left(\frac{b^{3/2}}{n}\right),
\end{align*}
where we use the bound: $1 - \sqrt{\zeta^b} = 1 - \zeta^{b/2} \geq \frac{b}{4n}.$
Thus, using the bound $\frac{1}{q'} = \frac{1}{\zeta^b} = O(1/e)$, we find that SMART reaches accuracy $\varepsilon$ in at most 
$O\left(\frac{2\eta}{q'\gamma\varepsilon} \right) = O\left(\frac{n}{b^{3/2}\varepsilon}\right)$ iterations. 

To initialize properly, SMART requires $n$ gradient evaluations. Then, on average, the $w$ variables are updated once every $1/(1- (1-1/n)^b) = O(n/b)$ steps, and each of those updates requires $n$ function evaluations, $n$ gradient evaluations, and $1$ evaluation of $\prox_{\tau r_1}$. Thus, to reach accuracy $\varepsilon$, SMART requires on average at most $O(n/(b^{3/2}\varepsilon)) n (b/n) = O(n/(b^{1/2}\varepsilon))$ function evaluations and $O(n/(b^{3/2}\varepsilon)) (b/n) = O(1/(b^{1/2}\varepsilon))$ evaluations of $\prox_{\tau r_1}$. 

Similarly, the $x$ variables are updated every $1/(1-1/n)^b = O(1)$ iterations, and each update requires $b$ gradient evaluations, and $1$ evaluation of $\prox_{\gamma r_2}$. Thus, to reach accuracy $\varepsilon$, SMART requires at most 
$$
n + O(n/(b^{3/2}\varepsilon)) n(b/n) +  O(n/(b^{3/2}\varepsilon)) b = O(n + n/(b^{1/2}\varepsilon))
$$ gradient evaluations and $O(n/(b^{3/2}\varepsilon))$ evaluations of $\prox_{\gamma r_2}$.

\section{Proof of Theorem~\ref{thm:convergencerate}}

We use the same notation from the proof of Theorem~\ref{thm:smartconverges} except that we redefine:
\begin{itemize}
\item \textbf{The $\delta(\kappa)$ Factor:} $\kappa \in (0, \left[\sqrt{q'(1-\rho_i)}\right]^{-1} - 1)$, we let
$$
\delta(\kappa) = \max_i\left\{1 - \mu\min\left\{\frac{q'\gamma}{2\eta}, \frac{q\tau}{2} \right\},  (1+\kappa)\sqrt{q'(1-\rho_i)}\right\}  \in (0, 1),
$$
\item \textbf{The $a$ Factor:}  
$$
a = \gamma\sqrt{\frac{1}{n}\sum_{i=1}^n   \frac{q' (1+\epsilon_0)(B_iL_{i})^2}{2 b\left(\sqrt{\delta(\kappa)} - \sqrt{q'(1-\rho_{i})}\right)^2}}. 
$$
\item \textbf{The $\beta_i$ Factor:} 
$$
\beta_{i} := \sqrt{1-\rho_{i}}\left(\frac{\sqrt{\delta(\kappa)}}{\sqrt{q'}} -\sqrt{1-\rho_{i}}\right).
$$
\item \textbf{The $\alpha_i$ Factor:}
%\begin{align*}
$$
\alpha_{i} :=  \frac{q'\gamma(1+\epsilon_0)}{2ab\left[ \delta(\kappa) - q'(1+\beta_{i})(1-\rho_{i})\right]}.
$$
%\end{align*}
Note that $\alpha_i$ is well-defined and positive because
$
q'(1+\beta_{i})(1-\rho_{i}) \leq q'(1+ \beta_i - \rho_i) = \sqrt{q'(1-\rho_i)} \leq \delta(\kappa)/(1+\kappa).
$
Then by definition of $\alpha_i$, we have
\begin{equation}\label{eq:linear_alpha}
\alpha_i q'(1+\beta_{i})(1-\rho_{i}) + \frac{q'\gamma(1+\epsilon_0)}{2ab} = \delta(\kappa) \alpha_i .
\end{equation}
\end{itemize}

With this choice of $\alpha_i$ the following bound holds (we defer the proof for a moment): 
\begin{lemma}[$\alpha_{i}$ bound]\label{lem:alpha_bound_linear}
The following bound holds: for all $k \in \NN$, $\kappa > 0$, we have
\begin{align*}
\frac{1}{n}\sum_{i=1}^n \left[\alpha_{i}(w_i^kL_i)^2\left( 1+ \frac{(1- \rho_{i} )}{\beta_{i}}\right)+ \frac{|w_i^k| L_i }{2}\right] \leq  \frac{1-2a}{2\gamma} .
\end{align*}
\end{lemma}

By an argument nearly identical to the argument in Theorem~\ref{thm:convergencerate} (recall~\eqref{eq:thing_to_cite_in_linear}), we have
\begin{align*}
&\EE_k\left[F(w^{k+1}, x^{k+1}) + \frac{1}{n}\sum_{i=1}^n \alpha_{i}  V^{k+1}_i \right] \leq F(w^k, x^k) + \frac{\delta(\kappa)}{n}\sum_{i=1}^n \alpha_{i}  \Dk_i - Y_k ,\numberthis\label{eq:lyapunov_error_bound}
\end{align*}
where $Y_k$ is defined in~\eqref{eq:supermartingale_definition}, and the properties of $\alpha_i$ defined in~\eqref{eq:linear_alpha} and Lemma~\ref{lem:alpha_bound_linear} play a key role. 
From the definition of $Y_k$ and the error bound~\eqref{eq:error_bound}, we find that
\begin{align*}
Y_k &\geq \frac{q'\gamma}{2\eta}\left\|\frac{\eta}{\gamma} \left(\overline{x}^{k+1} - x^k\right) \right\|^2 + \frac{q\tau}{2}\left\|\frac{1}{\tau}\left(\overline{w}^{k+1} - w^k\right)\right\|^2 \geq \min\left\{\frac{q'\gamma}{2\eta}, \frac{q\tau}{2} \right\} \mu \left[F(w^{k}, x^{k}) - F(w^\ast, x^\ast)\right]\\
&\geq (1-\delta(\kappa)) \mu \left[F(w^{k}, x^{k}) - F(w^\ast, x^\ast)\right].
\end{align*}
Thus, by plugging this bound into~\eqref{eq:lyapunov_error_bound}, we have
\begin{align*}
&\EE_k\left[F(w^{k+1}, x^{k+1}) - F(w^\ast, x^\ast) + \frac{1}{n}\sum_{i=1}^n \alpha_{i}  V^{k+1}_i \right] \leq \delta(\kappa)\left[F(w^k, x^k) - F(w^\ast, x^\ast) + \frac{1}{n}\sum_{i=1}^n \alpha_{i}  \Dk_i\right].
\end{align*}

To complete the proof, we use the law of total expectation to unfold the contraction: for all $k \in \NN$, we have 
\begin{align*}
&\EE\left[ F(w^{k+1}, x^{k+1}) - F(w^\ast, x^\ast) \right] \leq \EE\left[X_{k+1} - F(w^\ast, x^\ast)\right] \\
&\leq \delta(\kappa)\EE\left[X_k -F(w^\ast, x^\ast)\right] \leq \delta(\kappa)^{k+1}\left[X_0 - F(w^\ast, x^\ast)\right] = \delta(\kappa)^{k+1}\left[F(w^0, x^0) - F(w^\ast, x^\ast)\right].
\end{align*}
Take the limit as $\kappa \rightarrow 0$ to get the result.

\begin{proof}[of Lemma~\ref{lem:alpha_bound_linear}]
Let $i \in \{1, \ldots, n\}$ such that $\rho_i \neq 1$. Recall that 
$
\beta_{i} := \sqrt{1-\rho_{i}}\left(\frac{\sqrt{\delta(\kappa)}}{\sqrt{q'}} -\sqrt{1-\rho_{i}}\right).
$
Define numbers $\zeta_i := \sqrt{\delta(\kappa)} - \sqrt{q'(1-\rho_{i})}$ and $\theta_{i} := 1 - (1/\delta(\kappa))q'(1 + \beta_i- \rho_i) = 1 - (1/\sqrt{\delta(\kappa)})\sqrt{q'(1-\rho_i)}$, and note that 
%\begin{align*}
$
q'(1+\beta_{i})(1-\rho_{i} )/\delta(\kappa) \leq q'(1 + \beta_{i} - \rho_{i})/\delta(\kappa) =\sqrt{q'(1- \rho_i)}/\sqrt{\delta(\kappa)} =  1- \theta_{i}.
$
%\end{align*}
Thus, we have
%\begin{align*}
$$
\alpha_{i} \leq \frac{q'\gamma(1+\epsilon_0)}{2\delta(\kappa)ab} \sum_{t= 0}^{\infty} \left[(1-\theta_{i})\right]^t =  \frac{q'\gamma(1+\epsilon_0) }{2\delta(\kappa)ab}\frac{1}{\theta_i} \leq  \frac{q'\gamma(1+\epsilon_0) }{2ab\sqrt{\delta(\kappa)}\zeta_i}.
$$
%\end{align*}

With this bound in hand, we find that  
\begin{align*}
\alpha_{i} \left[ 1+ \frac{(1- \rho_{i})}{\beta_{i}}\right]  &\leq  \frac{q'\gamma(1+\epsilon_0) }{2ab\sqrt{\delta(\kappa)}\zeta_i}\left[  \frac{\sqrt{\delta(\kappa)}}{\zeta_i}\right]= \frac{q'\gamma(1+\epsilon_0) }{2ab\zeta_i^2}.
\end{align*}
Therefore, because
$
a = \gamma\sqrt{\frac{1}{n}\sum_{i=1}^n   \frac{q' (1+\epsilon_0)(B_iL_{i})^2}{2 b\zeta_i^2}},  
$
we have 
\begin{align*}
 &\frac{1}{n}\sum_{i=1}^n \left[q'\alpha_{i}(L_iw_i^k)^2\left( 1+ \frac{(1- \rho_{i} )}{\beta_{i}}\right)+ \frac{|w_i^k| L_i}{2}\right]  + \frac{a}{\gamma} 
 \\ &
 \leq\frac{1}{n}\left[\frac{\gamma}{a}\sum_{i=1}^n   \frac{q'(1+\epsilon_0) (B_i L_{i})^2}{2b \zeta_i^2}
+ \frac{B_i L_i }{2}\right]  + \frac{a}{\gamma} = 2\sqrt{\frac{1}{n}\sum_{i=1}^n   \frac{q'(1+\epsilon_0) (B_iL_{i})^2}{2b \zeta_i^2}} + \frac{1}{n}\sum_{i=1}^n \frac{B_i L_i }{2}\\% \leq \frac{1}{2\gamma} \\
&\hspace{80pt}\leq 2\sqrt{\frac{1}{n}\sum_{i=1}^n   \frac{q'(1+\epsilon_0) (B_iL_{i})^2}{2b \sqrt{q'(1-\rho_{i})}\left(1 - (q'(1-\rho_{i}))^{1/4}\right)^2}} + \frac{1}{n}\sum_{i=1}^n \frac{B_i L_i }{2} \leq \frac{1}{2\gamma},
\end{align*}
where the second to last line follows because 
$
\delta(\kappa) \geq (1+\kappa)\sqrt{q'(1-\rho_{i})} \geq \sqrt{q'(1-\rho_{i})},
$
and the last inequality holds by assumption.

\end{proof}

\subsection{Proof of Corollary~\ref{cor:linear_saga}}\label{app:linear_saga}

Our choice of $\tau$ guarantees that $\frac{q\tau}{2} = \frac{q'\gamma}{2\eta}$. Thus, from Theorem~\ref{thm:convergencerate}, it is clear that SMART achieves accuracy $\varepsilon$ after 
$
O\left(\log(1/\epsilon)/\log(1/\delta))\right)
$ 
iterations. We estimate this ratio below.

Because $D_k \equiv I_k$, we find that $\rho_i = P(i  \in D_k) = 1 - (1-1/n)^b$. Thus, with  
$$
a' = \sqrt{\frac{1}{n}\sum_{i=1}^n   \frac{q' (1+\epsilon_0)(B_iL_{i})^2}{2 b\sqrt{q'(1-\rho_{i})}\left(1 - (q'(1-\rho_{i}))^{1/4}\right)^2}},  
$$
and $\zeta = 1-1/n$ we have 
\begin{align*}
&\frac{\gamma}{2\eta} = \frac{1}{2(1/\gamma + 2a' + 2L\sum_{i=1}^nB_i)} =  \frac{1}{2(4 a' + \frac{L}{n}\sum_{i=1}^nB_i  + 2a' + \frac{2L}{n}\sum_{i=1}^nB_i)}\\
&= \frac{1}{2L\left( \frac{6\sqrt{\frac{1}{n}\sum_{i=1}^n q' (1+\epsilon_0)B_i^2}}{\sqrt{2b}\zeta^{(b+1)/4}\left(1 -\zeta^{(b+1)/4}\right)} + \frac{3}{n}\sum_{i=1}^n B_i\right) } \\
&= \frac{\sqrt{2b}\zeta^{(b+1)/4}\left(1 - \zeta^{(b+1)/4}\right)}{2L\left( 6\sqrt{\frac{1}{n}\sum_{i=1}^n q' (1+\epsilon_0)B_i^2} + \sqrt{2b}\zeta^{(b+1)/4}\left(1 - \zeta^{(b+1)/4}\right)\frac{3}{n} \sum_{i=1}^nB_i\right) } = \Omega\left(\frac{b^{3/2}}{Ln}\right),
\end{align*}
where we use the bounds: $1 - (1-1/n)^{(b+1)/2} \geq \frac{b+1}{8n}$ and $(1-1/n)^{(b+1)/4}  = \Omega (1/e).$ By definition, $\frac{1}{\log(1/\delta)}$ is smaller than the maximum of 
%\begin{align*}
\begin{align*}
  &\frac{1}{-\log\left(1 - \mu q' \frac{\gamma}{2\eta}\right)}\leq  \frac{1}{\mu q'}\frac{2\eta}{\gamma} =  O\left(\kappa \frac{n}{b^{3/2}}\right); \text{ and }\\
  &\frac{1}{-\log\left((q'(1-\rho_i))^{1/2}\right)} = \frac{1}{-\log\left(\zeta^{(b+1)/2}\right)} \leq \frac{1}{1-\zeta^{(b+1)/2}} = O\left(\frac{n}{b^{3/2}}\right).
\end{align*}
Therefore, SMART achieves accuracy $\varepsilon$ in at most 
$
\frac{\log(1/\varepsilon)}{\log(1/\delta)} = O\left(\kappa\frac{n}{b^{3/2}}\log(1/\epsilon)\right)
$
iterations. 

To initialize properly, SMART requires $n$ gradient evaluations. Then, on average, the $w$ variables will be updated once every $n$ steps, and each of those updates requires $n$ function evaluations, $n$ gradient evaluations, and evaluation of $\prox_{\tau r_1}$. Thus, to reach accuracy $\varepsilon$, SMART requires on average at most $O(\kappa(n/b^{3/2})\log(1/\varepsilon))$ function evaluations and $O(\kappa(1/b^{3/2})\log(1/\varepsilon))$ evaluations of $\prox_{\tau r_1}$. Similarly, the $x$ variables are updated every $1/(1-1/n) = O(1)$ iterations, and each update requires takes $b$ gradient evaluations, and $1$ evaluation of $\prox_{\gamma r_1}$. Thus, to reach accuracy $\varepsilon$, SMART requires at most $O(n + \kappa(n/b^{3/2} + n/b^{1/2})\log(1/\varepsilon))$ gradient evaluations and $O\left(\kappa(n/b^{3/2})\log(1/\varepsilon)\right)$ evaluations of $\prox_{\gamma r_2}$.

\end{document}